\documentclass{article} % For LaTeX2e
\usepackage[preprint]{colm2026_conference}

\usepackage{microtype}
\usepackage{hyperref}
\usepackage{url}
\usepackage{booktabs}
\usepackage{graphicx}
\usepackage{listings}

\usepackage{multirow}
\usepackage{amsmath}
\usepackage{amssymb}
\usepackage{xspace}
\usepackage{inconsolata} % optional, nicer monospace
\lstdefinestyle{promptstyle}{
  basicstyle=\ttfamily\footnotesize,
  breaklines=true,           % wrap long lines
  breakatwhitespace=false,   % wrap even inside long tokens if needed
  columns=fullflexible,      % better line breaking in two columns
  keepspaces=true,           % preserve spaces
  showstringspaces=false,
  frame=single,              % subtle box to visually separate blocks
  xleftmargin=0pt, xrightmargin=0pt,
  aboveskip=0.6em, belowskip=0.6em,
}

\usepackage{lineno}

\definecolor{darkblue}{rgb}{0, 0, 0.5}
\hypersetup{colorlinks=true, citecolor=darkblue, linkcolor=darkblue, urlcolor=darkblue}

\title{Single-Agent LLMs Outperform Multi-Agent Systems on\\ Multi-Hop Reasoning Under Equal Thinking Token Budgets}

\author{Dat Tran, Douwe Kiela\\
Stanford University\\
\texttt{\{dattran, dkiela\}@stanford.edu} \\
}

\begin{document}

\ifcolmsubmission
\linenumbers
\fi

\maketitle

\begin{abstract}
%Recent work suggests that single-agent LLM systems can match multi-agent systems when computation is normalized, but the theoretical underpinnings and methodological challenges of this comparison remain underexplored. We provide a formal information-theoretic argument via the Data Processing Inequality for why single-agent systems should hold an advantage under a fixed computation budget, and extend this perspective to clarify why multi-agent systems can become competitive when the single agent's effective context utilization is degraded. We then present a rigorous empirical study validating this theory, comparing single-agent systems and multiple multi-agent architectures under a fixed thinking-token budget across three model families (Qwen3, DeepSeek-R1-Distill-Llama, Gemini 2.5). Our results confirm that single-agent systems consistently match or outperform sequential and other multi-agent variants when thinking tokens are held constant, indicating that for multi-hop reasoning, single-agent systems are generally superior under comparable computation resources. More critically, our deep diagnostic analysis uncovers how these systems fail and reveals significant methodological artifacts in both API-based budgeting (Gemini 2.5) and standard benchmarks. Taken together, our results suggest that many reported multi-agent gains are better explained by unbudgeted compute and context effects than by inherent architectural superiority, while also identifying the degraded-context regimes in which multi-agent or hybrid designs are most promising.
Recent work reports strong performance from multi-agent LLM systems (MAS), but these gains are often confounded by increased test-time computation. When computation is normalized, single-agent systems (SAS) can match or outperform MAS, yet the theoretical basis and evaluation methodology behind this comparison remain unclear. We present an information-theoretic argument, grounded in the Data Processing Inequality, suggesting that under a fixed reasoning-token budget and with perfect context utilization, single-agent systems are more information-efficient. This perspective further predicts that multi-agent systems become competitive when a single agent’s effective context utilization is degraded, or when more compute is expended. We test these predictions in a controlled empirical study across three model families (Qwen3, DeepSeek-R1-Distill-Llama, and Gemini 2.5), comparing SAS with multiple MAS architectures under matched budgets. We find that SAS consistently match or outperform MAS on multi-hop reasoning tasks when reasoning tokens are held constant. Beyond aggregate performance, we conduct a detailed diagnostic analysis of system behavior and evaluation methodology. We identify significant artifacts in API-based budget control (particularly in Gemini 2.5) and in standard benchmarks, both of which can inflate apparent gains from MAS. Overall, our results suggest that, for multi-hop reasoning tasks, many reported advantages of multi-agent systems are better explained by unaccounted computation and context effects rather than inherent architectural benefits, and highlight the importance of understanding and explicitly controlling the trade-offs between compute, context, and coordination in agentic systems.
\end{abstract}

\section{Introduction}

Multi-agent LLM architectures (MAS), including planners, role-playing systems, debate frameworks, and tool-specialized swarms, have demonstrated strong empirical performance across a range of tasks. At a high level, these approaches decompose reasoning across multiple agents that operate over partial contexts and communicate via generated text. In contrast, single-agent systems (SAS) perform reasoning within a single, unified context, relying on internal token-level computation rather than explicit inter-agent communication.

However, comparisons between MAS and SAS are often confounded by differences in test-time computation. MAS typically consume more tokens through longer reasoning traces or multiple agent interactions, making it unclear whether their gains arise from architectural advantages or simply from increased compute. Recent budget-aware studies suggest that, when computation is normalized, many such strategies underperform strong single-agent baselines~\citep{wang2024reasoning,han2025token}.

%A central claim of this work is that \emph{under fixed token budgets, coordination trades off with information retention}. Intuitively, while MAS enable decomposition, they also introduce communication bottlenecks and fragmentation of context across agents.

In this work, we revisit this question under an explicit focus on \emph{thinking token budgets}, which we define as the total number of tokens used for intermediate reasoning, excluding prompts and final answers. We focus on multi-hop reasoning tasks and ask three central questions: \emph{why} might SAS outperform MAS under fixed budgets, \emph{when} do MAS become competitive, and \emph{how} should such comparisons be conducted reliably?

We first provide an information-theoretic perspective, based on the Data Processing Inequality, suggesting that under fixed token budgets, multi-agent decompositions introduce additional communication bottlenecks that can lead to information loss. This perspective also clarifies when MAS can be advantageous: specifically, when a single agent’s effective context utilization is degraded (e.g., due to long or noisy contexts), or when MAS benefit from additional unaccounted computation through extended interactions.

We then test these predictions in a controlled empirical study across three model families (Qwen3, DeepSeek-R1-Distill-Llama, and Gemini 2.5), comparing SAS and multiple MAS architectures under matched reasoning-token budgets. We find that SAS consistently match or outperform MAS on multi-hop reasoning tasks under these constraints.

Beyond aggregate performance, we perform a detailed diagnostic analysis of evaluation methodology and model behavior. We identify (i) artifacts in API-based budget control that distort effective computation, (ii) benchmark vulnerabilities exposed through paraphrasing, and (iii) systematic differences in failure modes across architectures.

We summarize our contributions as follows:

\begin{itemize}
\item An \textbf{information-theoretic perspective} on SAS vs.\ MAS under fixed budgets, explaining how multi-agent decompositions introduce additional communication bottlenecks (\S\ref{sec:theory_justification}).

\item A \textbf{controlled empirical comparison} across multiple model families, showing that SAS consistently match or outperform MAS under matched reasoning-token budgets on multi-hop reasoning tasks (\S\ref{sec:results}).

\item A \textbf{diagnostic analysis of evaluation methodology}, uncovering artifacts in API-based budgeting, benchmark vulnerabilities, and failure modes~(\S\ref{appendix:gemini_accounting}, \S\ref{appendix:paraphrase}, \S\ref{appendix:error_analysis}).

\end{itemize}

\section{Related Work}
% \paragraph{Budget-aware reasoning:}
% Reasoning in Token Economies (RTE) formalized token/compute-normalized comparisons and showed many complex strategies fail to beat self-consistency under matched budget~\citep{wang2024reasoning}. Follow-ups propose dynamic or learned budget allocation (TALE/Token-Budget-Aware)~\citep{han2025token} and budget-conditioned or RL-trained controllers (SelfBudgeter, BudgetThinker)~\citep{li2025selfbudgeter,wen2025budgetthinker}. Our work aligns with this line but isolates \emph{thinking tokens} as the controlled variable and evaluates MAS specifically.

\paragraph{Budget-Controlled Evaluation:}
Recent work has emphasized that comparisons between reasoning strategies are often confounded by unequal test-time computation. Reasoning in Token Economies and follow-up work show that many elaborate prompting or search strategies fail to outperform simpler baselines once token or compute budgets are matched~\citep{wang2024reasoning,han2025token,li2025selfbudgeter,wen2025budgetthinker}. Our work is aligned with this line, but differs in two ways: we treat only \emph{thinking tokens} as the controlled resource, and we study the comparison between single-agent and multi-agent architectures rather than budget allocation within one architecture.

\paragraph{When and why MAS helps:}
Concurrent works sharpen the picture that MAS gains are dependent on regime and implementation. \citet{anthropic2025researchsystem} argues that much of the apparent advantage of MAS comes from increased compute. On the other hand, \citet{kim2025towards} shows that once computation is normalized, agentic benefits tend to concentrate in weaker model or harder-regime settings and diminish (or even reverse) as base model capability increases, highlighting coordination overhead as a first-order constraint. Additionally, \citet{cemri2025multi} introduce multi-agent tracing and a failure taxonomy that explains how orchestration can induce drift, information loss, or misleading improvements from evaluation artifacts. These failure modes  align with our observed over-exploration, aggregation/extraction errors, and opaque API accounting. Finally, \citet{ke2026mas} study learned orchestration via holistic configuration selection and propose controlled benchmarking axes for MAS, emphasizing that improvements depend on task structure and verification protocols rather than MAS being universally superior. Taken together, these results motivate the core design of our study: (i) strictly control the primary confound (thinking token budget), (ii) diagnose how coordination and communication affect accuracy, and (iii) characterize concrete regimes (e.g., degraded context) where MAS or hybrid designs can become competitive.

\paragraph{SAS vs.\ MAS:}
Empirical evidence increasingly suggests that as frontier models improve, the benefits of orchestration diminish and SAS can match or surpass MAS~\citep{gao2025whynotboth}. In education analytics, SAS with few-shot prompting outperformed MAS for reflection assessment~\citep{li2025single}. Benchmarks for collaborative agents (e.g., MedAgentBoard) further reveal that MAS advantages are task-specific and not universal~\citep{zhu2025medagentboard}. Our findings complement these studies under stricter compute controls.

\paragraph{Multi-agent collaboration mechanisms: }
Prior work has proposed many ways to structure collaboration between LLMs, including debate, role specialization, ensemble/self-consistency, and reflection~\citep{du2024improving,zhang2023exploring,li2024more,shinn2023reflexion}. Rather than treating MAS as a single design, we evaluate several representative mechanisms under a common matched-budget framework. This allows us to separate gains due to architectural structure from gains due to simply spending more computation.

\paragraph{Context length, degradation, and long-context utilization:}
Independent of architecture, modern LLMs do not use long context perfectly: attention dilution, noise sensitivity, context confusion, and positional bias can all degrade reasoning even when relevant information is present. Recent work has formalized these effects from several angles, including context degradation and long-context stress testing~\citep{du2025context,li2025focusllm,lu2024controlled}, failures that are sensitive to their position such as the lost in the middle effect~\citep{liu2024lost}, and broader reliability decay as input length grows~\citep{hong2025contextrot}. We connect this literature to the SAS-versus-MAS question by explicitly modeling degraded effective context and by testing when structured multi-agent pipelines become competitive as single-agent context use deteriorates.

\section{Theoretical justification}
\label{sec:theory_justification}

This section aims to formalize the comparison between single-agent and multi-agent systems under a thinking-token budget, and explain both \emph{why} single agents often perform at least as well and \emph{when} multi-agent designs can help.

Let $Y$ denote the correct answer, $C$ the full context available to a single-agent LLM (including prior reasoning and intermediate states), and $M = g(C)$ the messages or summaries passed between agents in a multi-agent system.

Since $M$ is a (possibly stochastic) function of $C$, we have the Markov chain
\[
Y \longleftrightarrow C \longleftrightarrow M.
\]
By the \textbf{Data Processing Inequality (DPI)} \citep{CoverThomas2006},
\[
I(Y;C) \ge I(Y;M),
\]
and equivalently,
\[
H(Y \mid M) \ge H(Y \mid C).
\]
Thus, conditioning on $M$ leaves more residual uncertainty about $Y$ than conditioning on $C$; the multi-agent architecture cannot increase mutual information with the true answer.

% By \textbf{Fano’s Inequality} \citep{Fano1952}, for any predictor based on observations $X$,
% \[
% P_e(X) \ge \frac{H(Y \mid X) - 1}{\log(|\mathcal{Y}| - 1)},
% \]
% where $P_e(X)$ is the minimal achievable error probability. Combining with DPI gives
% \[
% P_e(M) \ge P_e(C).
% \]

Now we compare architectures through their \emph{minimum achievable prediction error}. For any observable representation $X$, let $\mathcal{D}_X$ denote the set of all possibly randomized estimators $\delta(\hat y \mid x)$, and define
\[
P_e(X)
\;=\;
\inf_{\delta \in \mathcal{D}_X}
\Pr[\hat Y \neq Y],
\qquad
\hat Y \sim \delta(\cdot \mid X).
\]
Thus, $P_e(X)$ is the smallest achievable error probability for a predictor that only observes $X$.

Since $M$ is generated from $C$ through a fixed channel $q(m \mid c)$, taking any estimator $\delta_M \in \mathcal{D}_M$ that predicts from $M$ would induce an estimator based on $C$ via
\[
\delta_C^{\delta_M}(\hat y \mid c)
=
\sum_m q(m \mid c)\,\delta_M(\hat y \mid m).
\]
This induced estimator first samples the same message $M$ that the multi-agent pipeline would generate from $C$, and then applies the same downstream decision rule $\delta_M$. Therefore it reproduces the same joint distribution over $(Y,\hat Y)$ as the original predictor based on $M$:
\[
\Pr[\hat Y_{\delta_C^{\delta_M}} \neq Y]
=
\Pr[\hat Y_{\delta_M} \neq Y].
\]
Since every such induced estimator belongs to $\mathcal{D}_C$,
\[
\begin{aligned}
P_e(C)
&=
\inf_{\delta \in \mathcal{D}_C}
\Pr[\hat Y_\delta \neq Y]
\\
&\le
\inf_{\delta_M \in \mathcal{D}_M}
\Pr[\hat Y_{\delta_C^{\delta_M}} \neq Y]
\\
&=
\inf_{\delta_M \in \mathcal{D}_M}
\Pr[\hat Y_{\delta_M} \neq Y]
\\
&=
P_e(M).
\end{aligned}
\]

Hence, the single-agent system (with full access to $C$) is information-theoretically guaranteed to perform at least as well as the multi-agent system operating on $M = g(C)$.

\subsection{Context degradation}

In practice, however, modern LLMs may not utilize all of \(C\) equally well. To model this, let \(\tilde{C}_{\alpha} = T_{\alpha}(C)\) denote the \emph{effective context} available to the single-agent predictor under degradation level \(\alpha \ge 0\), where \(T_{\alpha}\) may represent deletion, masking, substitution noise, distractor injection, or more generally any transformation that makes relevant information harder to recover. We assume \(T_{0}\) is the identity map and that degradation is monotone in the sense that for \(0 \le \alpha_1 \le \alpha_2\), there exists a stochastic map \(S_{\alpha_1 \to \alpha_2}\) such that
\[
\tilde{C}_{\alpha_2} = S_{\alpha_1 \to \alpha_2}(\tilde{C}_{\alpha_1}).
\]
Then we obtain the Markov chain
\[
Y \longleftrightarrow C \longleftrightarrow \tilde{C}_{\alpha_1} \longleftrightarrow \tilde{C}_{\alpha_2},
\]
and therefore, by the Data Processing Inequality,
\[
I(Y;\tilde{C}_{\alpha_1}) \ge I(Y;\tilde{C}_{\alpha_2}).
\]
Equivalently, residual uncertainty is non-decreasing with degradation:
\[
H(Y \mid \tilde{C}_{\alpha_1}) \le H(Y \mid \tilde{C}_{\alpha_2}).
\]
The same argument as above implies that heavier degradation cannot improve the best achievable prediction error. Hence
\[
P_e(\tilde{C}_{\alpha_1}) \le P_e(\tilde{C}_{\alpha_2}).
\]

Suppose a multi-agent architecture constructs a collection of intermediate messages
\[
M_{\alpha} = g_{\alpha}(C),
\]
where \(g_{\alpha}\) may correspond to a structured transformation of the original context. The original DPI comparison only guarantees \(I(Y;C) \ge I(Y;M_{\alpha})\), not \(I(Y;\tilde{C}_{\alpha}) \ge I(Y;M_{\alpha})\). In other words, once effective single-agent context utilization deteriorates sufficiently, a well-designed MAS may recover task-relevant information more reliably than a degraded single pass, even though it still cannot exceed the ideal information available in \(C\).

This yields a concrete prediction for our experiments. Under low degradation, SAS should dominate because it retains access to the richest available representation of the task. As degradation increases, the SAS advantage may shrink, since the practical single-agent predictor now operates on an increasingly lossy effective context. In sufficiently degraded regimes, MAS and SAS may become comparable, and carefully structured MAS pipelines may occasionally surpass SAS by imposing useful factorization, filtering, or verification structure on the reasoning process.

\section{Approach}
\subsection{Tasks and Datasets}
We evaluate on \textbf{FRAMES} \citep{krishna2025fact} and \textbf{MuSiQue} \citep{trivedi2022musique}, both are multi-hop world knowledge questions with concise ground truth answers. For MuSiQue we filter to only 4-hop questions, which are complex enough to challenge both SAS and MAS. We report accuracy via an LLM-as-judge rubric checking whether the ground truth appears or is semantically present in the model answer (see \S\ref{sec:metric}).

\subsection{Single-agent architecture}

Our SAS pipeline is a single, direct pass. For a given question, the model receives a system prompt instructing it to ``think step by step, then answer''. It is allocated the \emph{entire} global thinking budget $B$ for a single call. The final answer is then extracted by taking the text that follows the \texttt{</think>} tag or Gemini API call.

\paragraph{SAS with Longer Thinking Variant (SAS-L).}
In \S\ref{appendix:gemini_accounting} we show that, for Gemini-2.5 (Flash/Pro), the \emph{visible} thought text produced by SAS tends to plateau well below the requested budget, while MAS surfaces more visible thought content under the same requested budget \(B\), due to multiple calls. To make SAS produce more thinking output, we introduce a SAS variant that gently encourages more internal reasoning before answering, without changing budget \(B\) or any other compute knobs.

The variant keeps the same single-call SAS pipeline and the same requested thinking budget \(B\) but augments the user message with a short, structured pre-answer analysis scaffold. Concretely, we add an extra instruction that asks the model to (i) identify ambiguities, (ii) propose interpretations, (iii) evaluate and select one, and only then (iv) answer.

\subsection{Multi-agent Architectures}

Our study includes one primary multi-agent baseline, \textbf{Sequential}, together with four additional multi-agent variants: \textbf{Subtask-parallel}, \textbf{Parallel-roles}, \textbf{Debate}, and \textbf{Ensemble}. All architectures operate under the same global thinking-token budget \(B\), with planner and aggregator components kept near budget-neutral whenever possible.

Our full multi-agent family is as follows:

\begin{enumerate}
    \item \textbf{Sequential:} A planner decomposes the question into ordered steps, the budget is split across sequential workers, and an aggregator synthesizes the intermediate outputs.

    \item \textbf{Subtask-parallel:} A planner proposes a small set of approximately independent subtasks, workers solve them in parallel under equal budget splits, and an aggregator combines their outputs.

    \item \textbf{Parallel-roles:} The full question is sent to role-specialized workers, Solver, Fact Extractor, Skeptic, and Second Solver. The Solver and Second Solver independently attempt to answer the question directly; the Fact Extractor identifies the key entities, constraints, and intermediate facts; and the Skeptic highlights possible pitfalls or alternative interpretations. The total budget \(B\) is divided evenly across these roles. At the end, an aggregator synthesizes their outputs.

    \item \textbf{Debate:} Two debaters first answer independently, then critique one another. The total budget \(B\) is divided evenly across these roles. Finally, an aggregator returns the final answer from the candidate answers and critiques.

    \item \textbf{Ensemble:} Multiple workers answer independently under equal budget splits with higher sampling temperature, and a judge selects the best candidate answer.
\end{enumerate}

\begin{figure}[h]
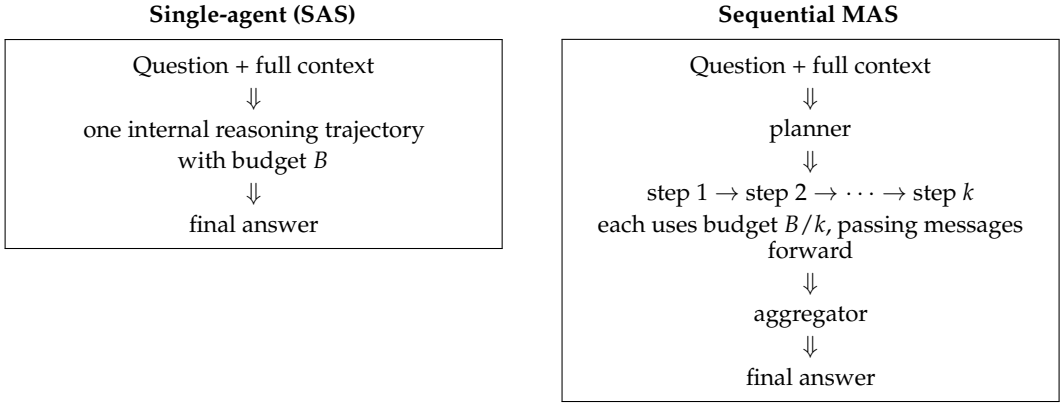

\centering
\setlength{\fboxsep}{6pt}
\small
\begin{minipage}[t]{0.47\linewidth}
\centering
\textbf{Single-agent (SAS)}\\[3pt]
\fbox{\parbox{0.94\linewidth}{
\centering
Question + full context\\[2pt]
\(\Downarrow\)\\[2pt]
one internal reasoning trajectory\\[2pt]
with budget \(B\)\\[2pt]
\(\Downarrow\)\\[2pt]
final answer
}}
\end{minipage}
\hfill
\begin{minipage}[t]{0.47\linewidth}
\centering
\textbf{Sequential MAS}\\[3pt]
\fbox{\parbox{0.94\linewidth}{
\centering
Question + full context\\[2pt]
\(\Downarrow\)\\[2pt]
planner\\[2pt]
\(\Downarrow\)\\[2pt]
step 1 \(\rightarrow\) step 2 \(\rightarrow \cdots \rightarrow\) step \(k\)\\[2pt]
each uses budget \(B/k\), passing messages forward\\[2pt]
\(\Downarrow\)\\[2pt]
aggregator\\[2pt]
\(\Downarrow\)\\[2pt]
final answer
}}
\end{minipage}
\caption{Sequential is the closest MAS comparator to SAS. Both are serial reasoning pipelines over the same question and the same global budget. The essential difference is that Sequential MAS externalizes intermediate reasoning into explicit messages between steps, whereas SAS keeps those intermediates latent within one continuous reasoning trajectory.}
\label{fig:sas-vs-seq}
\end{figure}

We obtain results for all of these multi-agent architectures and use \textbf{Sequential} as the main comparison against SAS for deeper analysis, as it is the cleanest multi-agent analogue of single-agent reasoning. Both SAS and Sequential attempt to solve the entire question through a serial reasoning process over one evolving trajectory. The key difference is that SAS keeps intermediate reasoning latent within a single chain, while Sequential externalizes intermediate states as explicit messages passed between steps. This makes Sequential the most appropriate architecture for the paper's core SAS-versus-MAS comparison: it isolates the cost and benefit of message-passing without simultaneously adding large amounts of diversity, specialization, or adversarial interaction. See Figure \ref{fig:sas-vs-seq}.

\subsection{Evaluation Metric}
\label{sec:metric}

We follow the standard \emph{LLM-as-a-judge} paradigm, in which a separate evaluation model scores candidate answers according to a fixed natural-language rubric. Concretely, for each example we provide the judge with the question, the model's prediction, and the ground-truth answer, together with a short decision rubric that specifies what counts as a correct answer (semantic equivalence, allowance for paraphrases, handling of minor formatting differences, etc.). The judge's rubric and prompt are used for all datasets and for both SAS and MAS configurations, so any differences in accuracy arise from the systems being evaluated, not from the judge.\footnote{Our design is inspired by the rubric-based evaluation setup in~\citet{krishna2025fact}, adapted to our open-ended answers.}

\begin{table*}[t]
\centering
\caption{Single-agent (SAS) vs.\ multi-agent architectures under matched \emph{thinking-token} budgets on FRAMES and MuSiQue. SAS-L is the ``longer thinking'' single-agent variant. ``Seq'' denotes our primary Sequential multi-agent baseline. ``Sub'', ``Roles'', ``Deb'', and ``Ens'' denote subtask-parallel, parallel roles, debate, and ensemble architectures, respectively. Bold indicates the highest-accuracy systems and every other system whose 95\% bootstrap confidence interval overlaps with that highest system's confidence interval (see Appendix \ref{appendix:confidence} for bootstrap confidence interval values).}
\label{tab:main-sas-mas}
\resizebox{\linewidth}{!}{%
\begin{tabular}{ll|ccccccc|ccccccc}
\toprule
& & \multicolumn{7}{c|}{\textbf{100 thinking tokens}} & \multicolumn{7}{c}{\textbf{500 thinking tokens}} \\
\cmidrule(lr){3-9}\cmidrule(lr){10-16}
\textbf{Data} & \textbf{Model} &
SAS & SAS-L & Seq & Sub & Roles & Deb & Ens &
SAS & SAS-L & Seq & Sub & Roles & Deb & Ens \\
\midrule
FRAMES & Qwen3-30B
& \textbf{0.191} & \textbf{0.195} & \textbf{0.198} & 0.155 & \textbf{0.207} & \textbf{0.204} & 0.146
& \textbf{0.240} & \textbf{0.220} & 0.223 & 0.187 & 0.223 & 0.206 & 0.193 \\
MuSiQue & Qwen3-30B
& \textbf{0.200} & \textbf{0.210} & \textbf{0.220} & 0.174 & 0.202 & \textbf{0.225} & 0.149
& \textbf{0.250} & 0.213 & 0.226 & 0.187 & 0.204 & 0.229 & 0.183 \\
\midrule
FRAMES & DeepSeek-R1-70B
& 0.365 & 0.374 & 0.387 & 0.385 & \textbf{0.427} & \textbf{0.400} & 0.365
& \textbf{0.427} & \textbf{0.432} & 0.396 & 0.402 & 0.407 & 0.392 & 0.400 \\
MuSiQue & DeepSeek-R1-70B
& 0.294 & \textbf{0.316} & \textbf{0.328} & \textbf{0.309} & \textbf{0.316} & 0.308 & \textbf{0.316}
& \textbf{0.383} & \textbf{0.375} & 0.332 & 0.303 & 0.329 & 0.320 & 0.319 \\
\midrule
FRAMES & G2.5-F
& 0.333 & \textbf{0.427} & \textbf{0.462} & 0.365 & \textbf{0.441} & \textbf{0.445} & 0.284
& \textbf{0.487} & \textbf{0.459} & \textbf{0.494} & 0.442 & 0.451 & \textbf{0.470} & 0.384 \\
MuSiQue & G2.5-F
& 0.263 & \textbf{0.352} & 0.272 & 0.264 & 0.300 & 0.318 & 0.243
& \textbf{0.340} & \textbf{0.352} & 0.287 & 0.291 & \textbf{0.311} & \textbf{0.329} & 0.274 \\
\midrule
FRAMES & G2.5-P
& 0.368 & 0.451 & \textbf{0.654} & 0.530 & 0.609 & \textbf{0.649} & 0.407
& 0.600 & 0.450 & \textbf{0.660} & 0.562 & 0.598 & \textbf{0.660} & 0.400 \\
MuSiQue & G2.5-P
& 0.308 & 0.373 & \textbf{0.393} & 0.364 & \textbf{0.400} & \textbf{0.412} & 0.302
& 0.391 & \textbf{0.423} & 0.391 & 0.363 & 0.397 & \textbf{0.435} & 0.330 \\
\midrule
Average & --
& 0.290 & 0.337 & \textbf{0.364} & 0.322 & \textbf{0.363} & \textbf{0.370} & 0.280
& \textbf{0.390} & 0.366 & \textbf{0.376} & 0.342 & 0.365 & \textbf{0.380} & 0.310 \\
\midrule
& & \multicolumn{7}{c|}{\textbf{1000 thinking tokens}} & \multicolumn{7}{c}{\textbf{2000 thinking tokens}} \\
\cmidrule(lr){3-9}\cmidrule(lr){10-16}
\textbf{Data} & \textbf{Model} &
SAS & SAS-L & Seq & Sub & Roles & Deb & Ens &
SAS & SAS-L & Seq & Sub & Roles & Deb & Ens \\
\midrule
FRAMES & Qwen3-30B
& \textbf{0.252} & \textbf{0.235} & 0.225 & 0.220 & 0.240 & 0.228 & 0.210
& \textbf{0.250} & \textbf{0.246} & \textbf{0.252} & \textbf{0.232} & \textbf{0.256} & \textbf{0.232} & \textbf{0.230} \\
MuSiQue & Qwen3-30B
& \textbf{0.260} & 0.231 & 0.229 & 0.207 & 0.220 & 0.224 & 0.197
& \textbf{0.271} & 0.239 & 0.229 & 0.234 & 0.226 & 0.234 & 0.224 \\
\midrule
FRAMES & DeepSeek-R1-70B
& \textbf{0.448} & \textbf{0.448} & 0.391 & \textbf{0.420} & \textbf{0.425} & 0.397 & \textbf{0.419}
& \textbf{0.454} & \textbf{0.451} & 0.393 & \textbf{0.434} & \textbf{0.433} & \textbf{0.425} & \textbf{0.431} \\
MuSiQue & DeepSeek-R1-70B
& \textbf{0.407} & \textbf{0.402} & 0.320 & 0.317 & 0.334 & 0.315 & 0.323
& \textbf{0.418} & \textbf{0.411} & 0.327 & 0.317 & 0.335 & 0.352 & 0.346 \\
\midrule
FRAMES & G2.5-F
& \textbf{0.551} & 0.484 & 0.507 & 0.482 & 0.495 & \textbf{0.527} & 0.458
& \textbf{0.532} & \textbf{0.517} & \textbf{0.526} & \textbf{0.509} & \textbf{0.536} & \textbf{0.527} & 0.498 \\
MuSiQue & G2.5-F
& \textbf{0.331} & \textbf{0.354} & 0.287 & 0.313 & 0.306 & \textbf{0.328} & 0.299
& 0.334 & \textbf{0.369} & 0.283 & 0.323 & 0.330 & \textbf{0.349} & 0.320 \\
\midrule
FRAMES & G2.5-P
& \textbf{0.680} & 0.610 & \textbf{0.670} & 0.596 & 0.600 & \textbf{0.638} & 0.430
& \textbf{0.700} & \textbf{0.680} & \textbf{0.690} & 0.637 & \textbf{0.658} & 0.660 & 0.558 \\
MuSiQue & G2.5-P
& 0.413 & 0.414 & 0.400 & 0.397 & \textbf{0.430} & \textbf{0.444} & 0.325
& 0.412 & \textbf{0.449} & 0.414 & 0.381 & 0.409 & \textbf{0.448} & 0.372 \\
\midrule
Average & --
& \textbf{0.418} & 0.397 & 0.379 & 0.369 & 0.381 & 0.388 & 0.333
& \textbf{0.421} & \textbf{0.420} & 0.389 & 0.383 & 0.398 & \textbf{0.403} & 0.372 \\
\midrule
& & \multicolumn{7}{c|}{\textbf{5000 thinking tokens}} & \multicolumn{7}{c}{\textbf{10000 thinking tokens}} \\
\cmidrule(lr){3-9}\cmidrule(lr){10-16}
\textbf{Data} & \textbf{Model} &
SAS & SAS-L & Seq & Sub & Roles & Deb & Ens &
SAS & SAS-L & Seq & Sub & Roles & Deb & Ens \\
\midrule
FRAMES & Qwen3-30B
& \textbf{0.260} & \textbf{0.246} & \textbf{0.252} & 0.237 & \textbf{0.265} & \textbf{0.238} & \textbf{0.254}
& \textbf{0.263} & \textbf{0.246} & \textbf{0.258} & \textbf{0.244} & \textbf{0.271} & 0.240 & \textbf{0.263} \\
MuSiQue & Qwen3-30B
& \textbf{0.271} & \textbf{0.248} & 0.231 & 0.249 & 0.246 & 0.249 & 0.226
& \textbf{0.271} & \textbf{0.248} & 0.231 & 0.254 & 0.242 & 0.244 & 0.245 \\
\midrule
FRAMES & DeepSeek-R1-70B
& \textbf{0.455} & \textbf{0.444} & 0.394 & \textbf{0.436} & \textbf{0.448} & \textbf{0.439} & \textbf{0.450}
& \textbf{0.456} & \textbf{0.445} & 0.397 & 0.434 & \textbf{0.450} & \textbf{0.444} & \textbf{0.458} \\
MuSiQue & DeepSeek-R1-70B
& \textbf{0.419} & \textbf{0.412} & 0.323 & 0.317 & 0.345 & 0.357 & 0.334
& \textbf{0.417} & \textbf{0.412} & 0.327 & 0.319 & 0.354 & 0.360 & 0.330 \\
\midrule
FRAMES & G2.5-F
& \textbf{0.545} & \textbf{0.542} & \textbf{0.524} & \textbf{0.532} & \textbf{0.545} & \textbf{0.552} & \textbf{0.539}
& \textbf{0.547} & \textbf{0.546} & \textbf{0.516} & \textbf{0.541} & \textbf{0.556} & \textbf{0.564} & \textbf{0.559} \\
MuSiQue & G2.5-F
& \textbf{0.344} & \textbf{0.364} & 0.289 & 0.332 & 0.340 & \textbf{0.355} & 0.338
& \textbf{0.338} & \textbf{0.369} & 0.285 & 0.335 & \textbf{0.349} & \textbf{0.354} & \textbf{0.343} \\
\midrule
FRAMES & G2.5-P
& \textbf{0.700} & \textbf{0.690} & 0.680 & 0.652 & \textbf{0.700} & \textbf{0.700} & \textbf{0.710}
& \textbf{0.692} & \textbf{0.692} & \textbf{0.691} & 0.655 & \textbf{0.718} & \textbf{0.697} & \textbf{0.719} \\
MuSiQue & G2.5-P
& 0.419 & \textbf{0.455} & 0.392 & 0.417 & \textbf{0.447} & \textbf{0.470} & 0.434
& 0.428 & 0.436 & 0.392 & 0.410 & \textbf{0.447} & \textbf{0.458} & 0.445 \\
\midrule
Average & --
& \textbf{0.427} & \textbf{0.425} & 0.386 & 0.396 & \textbf{0.417} & \textbf{0.420} & \textbf{0.411}
& \textbf{0.426} & \textbf{0.424} & 0.387 & 0.399 & \textbf{0.423} & \textbf{0.420} & \textbf{0.420} \\
\bottomrule
\end{tabular}
}
\end{table*}

\section{Results}
\label{sec:results}

Table~\ref{tab:main-sas-mas} summarizes our core comparison between SAS and multiple multi-agent architectures under matched thinking-token budgets across all models and datasets.
We report FRAMES and MuSiQue 4-hop for Qwen3-30B-A3B and DeepSeek-R1-Distill-Llama-70B, Gemini-2.5-Flash and Gemini-2.5-Pro.
\subsection{SAS vs.\ Multi-Agent Results Across Models and Datasets}
\label{subsec:main-main-results}

Our main finding is that, \textbf{under matched thinking token budgets} (except for very small budgets, which essentially do not produce any reasoning), \textbf{SAS is the strongest default architecture for multi-hop reasoning}. 
Across model families and datasets, SAS is the best-performing system or statistically indistinguishable from the best for all budgets except the lowest one (100 tokens), in which case the model does not produce a useful reasoning trace at all (for either approach). \textbf{SAS also consumes much less thinking token than any MAS variants while achieving the same or better results} (refer to Appendix \ref{appendix:confidence} for full results with thinking token spent).

\paragraph{SAS-L improves accuracy mostly for Gemini models.}
On Gemini-2.5-Flash, SAS-L is the strongest single-agent configuration in every MuSiQue panel, and its block averages remain above standard SAS throughout.
On Gemini-2.5-Pro MuSiQue, SAS-L is also generally stronger than standard SAS, especially in higher thinking token budgets.
%This supports the interpretation from the accounting section: for Gemini, prompting the model to better utilize its thinking channel can matter substantially.
By contrast, SAS-L is not a universal gain in the open-source setting.
For Qwen3-30B-A3B, SAS-L is sometimes competitive in the low-budget panel, but standard SAS is usually stronger in the middle and high-budget ranges.
%DeepSeek does not have SAS-L runs in the updated table.
In conclusion, SAS-L is mainly a Gemini effect: it helps when the Gemini models' thinking channel appears underutilized, but it does not produce the same consistent benefit for the open-source runs.

\textbf{Debate is the most consistently strong MAS variant.}
It repeatedly appears in the highest-confidence range, and on Gemini-2.5-Pro MuSiQue it is often the strongest overall MAS architecture.
Parallel-roles is frequently the next strongest and is especially competitive on FRAMES and on Gemini-Pro settings. Ensemble is more dependent on thinking token budgets.
It is usually weaker than Debate and often weaker than Parallel-roles on MuSiQue, especially at low and medium budgets.
However, on Gemini-2.5-Pro FRAMES at high budgets, Ensemble becomes strong and even attains the best accuracies in the 5000 and 10000 thinking token budgets.

\textbf{Performance improves as the thinking token budget increases and then flattens out.}
For many models/architectures, the 1000/2000 thinking token average is already close to the 5000/10000 thinking token average, even when the accuracies continue to fluctuate slightly. Although the trend is not perfectly monotonic, the results suggest diminishing returns from simply increasing the thinking budget beyond a certain point.
%
%But the broad pattern is consistent: \textbf{more thinking budget helps most in moving systems out of the extremely budget-constrained regime, after which gains become small and varies between architectures.}
At higher thinking token budgets, the models begin to saturate, and in some cases may over-explore or overthink rather than convert extra budget into better final answers.

Other observations include that \textbf{Gemini-2.5-Pro is the strongest overall} and \textbf{MuSiQue is clearly the harder benchmark.} DeepSeek-R1-Distill-Llama-70B is the stronger of the two open-source families. For MuSiQue, the confidence intervals also tend to be wider or more overlapping than FRAMES, reflecting greater instability and a harder reasoning problem.

\begin{figure}
    \centering
    \includegraphics[width=0.75\linewidth]{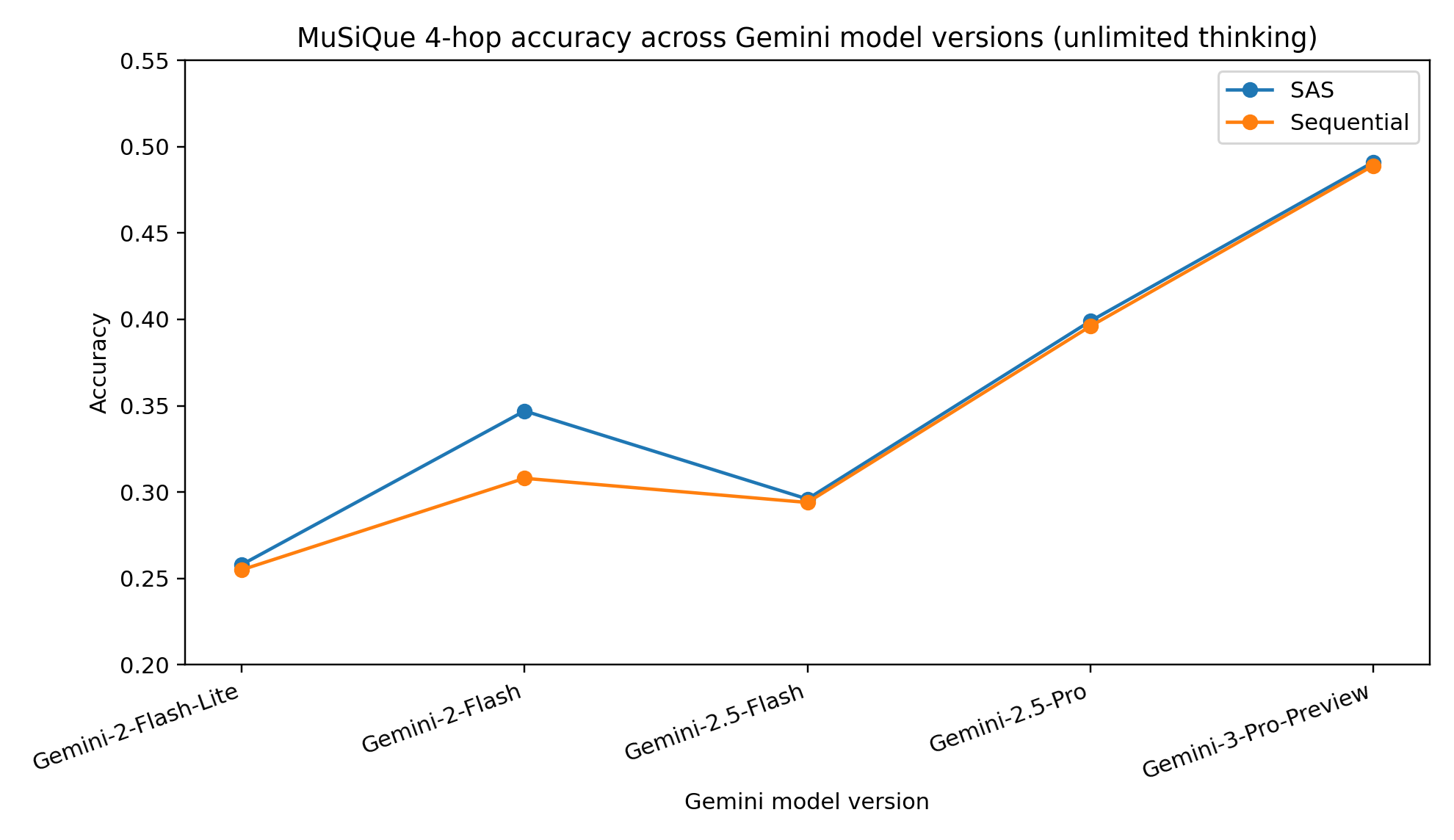}
    \caption{MuSiQue 4-hop accuracy across Gemini model versions with unlimited thinking
tokens.}
    \label{fig:gemini_multi_ver}
\end{figure}

\subsection{Comparing Gemini model versions}
\label{subsec:results-model-versions}
% To ensure that our conclusions are not artifacts of a particular model release, we swept across several Gemini model versions on MuSiQue 4-hop with \emph{unlimited} thinking tokens (dynamic thinking mode).
% Figure~\ref{fig:gemini_multi_ver} visualizes the full pattern across all Gemini models.

% Two patterns emerge.
% First, accuracy increases monotonically with model capability for both SAS and Sequential MAS: Gemini-2-Flash-Lite $<$ Gemini-2-Flash $<$ Gemini-2.5-Flash $<$ Gemini-2.5-Pro $<$ Gemini-3-Pro-Preview.
% For example, SAS accuracy rises from 0.258 (Gemini-2-Flash-Lite) to 0.347 (Gemini-2-Flash) to 0.399 (Gemini-2.5-Pro) to 0.491 (Gemini-3-Pro-Preview), with Sequential MAS tracking closely but slightly below. 
% Second, even under unlimited thinking, SAS remains competitive or marginally better than Sequential MAS for all versions; differences are within confidence intervals for the strongest model (Gemini-3-Pro-Preview).

% These comparisons suggest that our evaluation protocol is aligned with model quality (more capable models achieve higher accuracy) and that the SAS $\ge$ Sequential MAS pattern is not an artifact of a particular Gemini release.

To test whether the main result depends on a particular API release, we sweep several Gemini model versions on MuSiQue using unlimited thinking tokens. Figure~\ref{fig:gemini_multi_ver} shows two stable patterns. First, starting from Gemini-2.5-Flash, \textbf{performance increases monotonically with model capability for both SAS and Sequential MAS}. This also indicates that the evaluation is aligned with model quality as opposed to e.g. noise. Second, \textbf{SAS remains competitive with, and usually slightly stronger than, Sequential throughout the sweep.}

%For example, SAS accuracy rises from 0.258 on Gemini-2-Flash-Lite to 0.347 on Gemini-2-Flash, 0.399 on Gemini-2.5-Pro, and 0.491 on Gemini-3-Pro-Preview, with Sequential tracking closely but generally below.
Thus, even when we remove the explicit token cap and move across multiple Gemini generations, the same qualitative pattern persists: stronger models improve both architectures, but they do not create a regime in which Sequential becomes systematically superior.
These results strengthen the interpretation of Table \ref{tab:main-sas-mas}. The SAS $\ge$ Sequential pattern is not an artifact of one specific Gemini checkpoint or release window. Instead, it appears to be a stable property of the comparison itself, with both architectures benefiting from stronger base models but SAS remaining at least as competitive as the best multi-agent analogue.

\begin{figure*}[t]
    \centering

    \begin{minipage}[t]{0.48\textwidth}
        \centering
        \includegraphics[width=\linewidth]{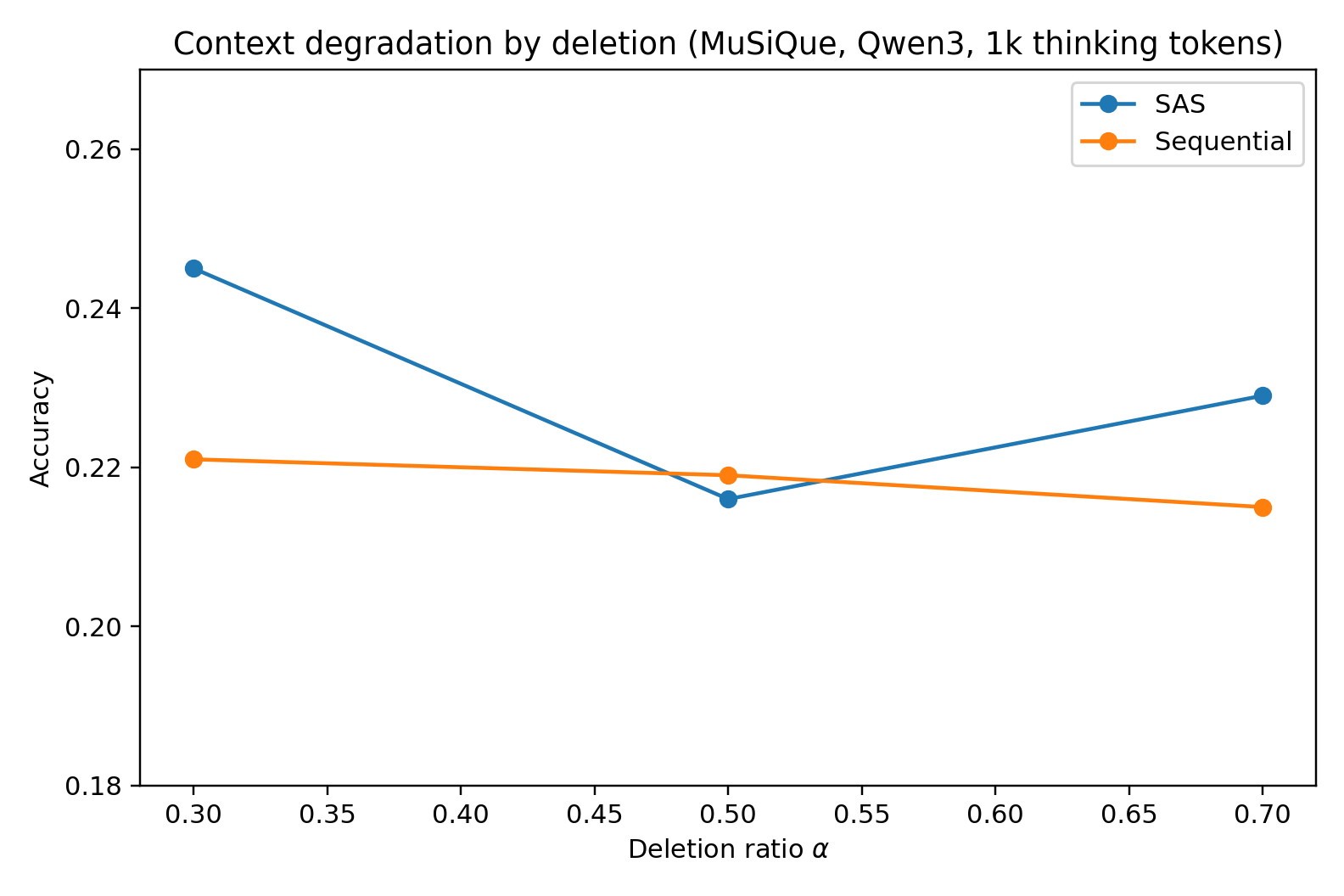}
    \end{minipage}
    \begin{minipage}[t]{0.48\textwidth}
        \centering
        \includegraphics[width=\linewidth]{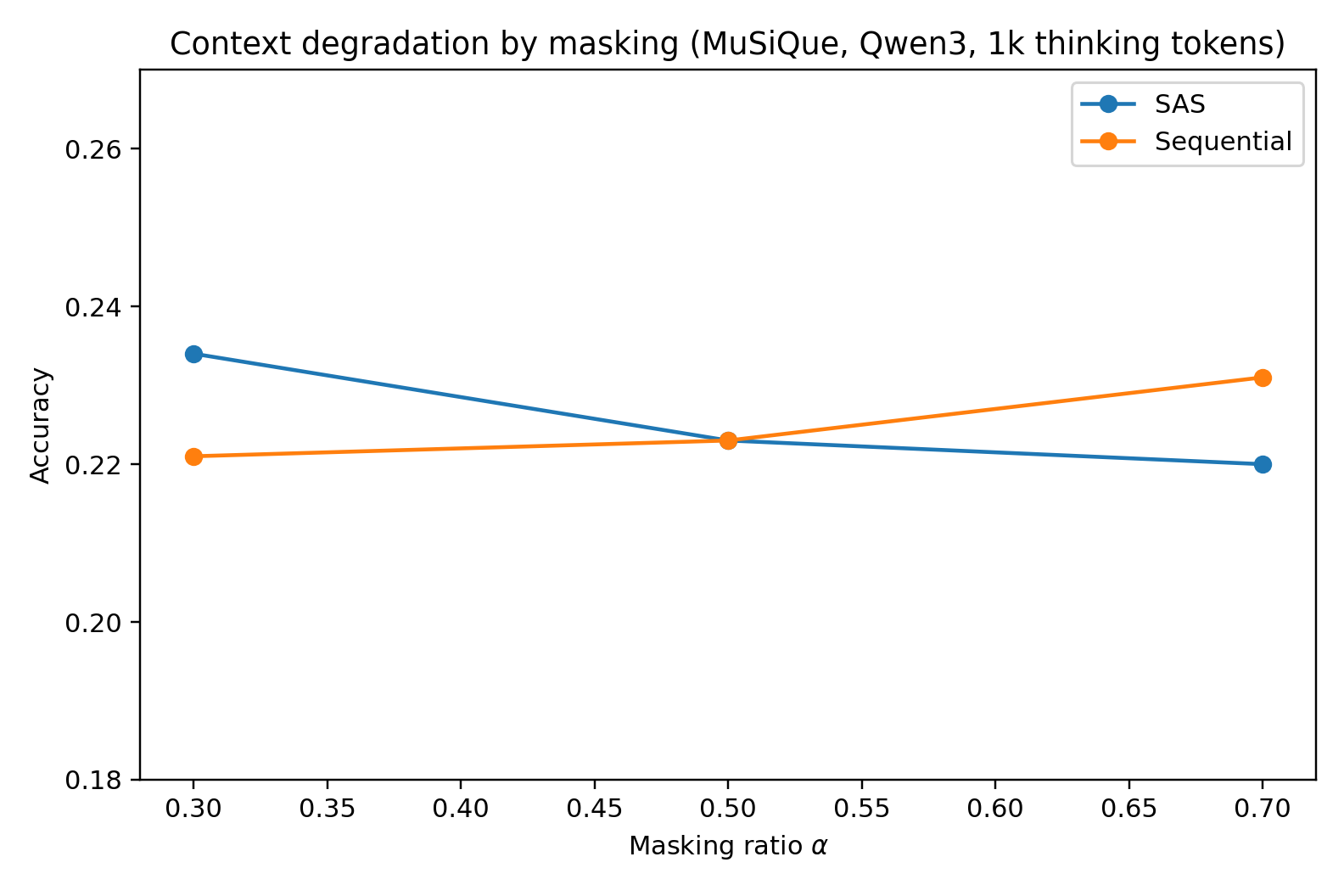}
    \end{minipage}
    \begin{minipage}[t]{0.48\textwidth}
        \centering
        \includegraphics[width=\linewidth]{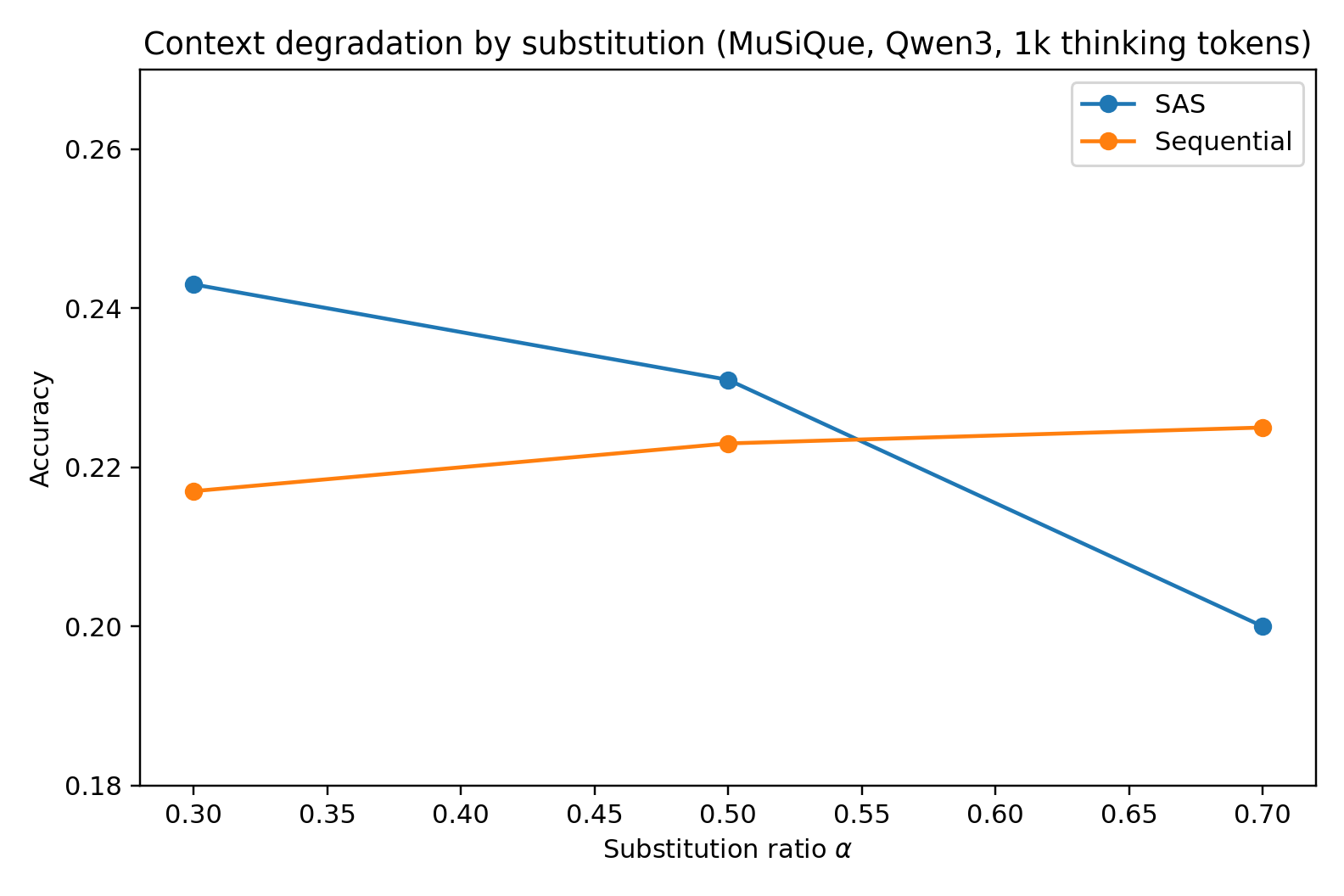}
    \end{minipage}
    \begin{minipage}[t]{0.48\textwidth}
        \centering
        \includegraphics[width=\linewidth]{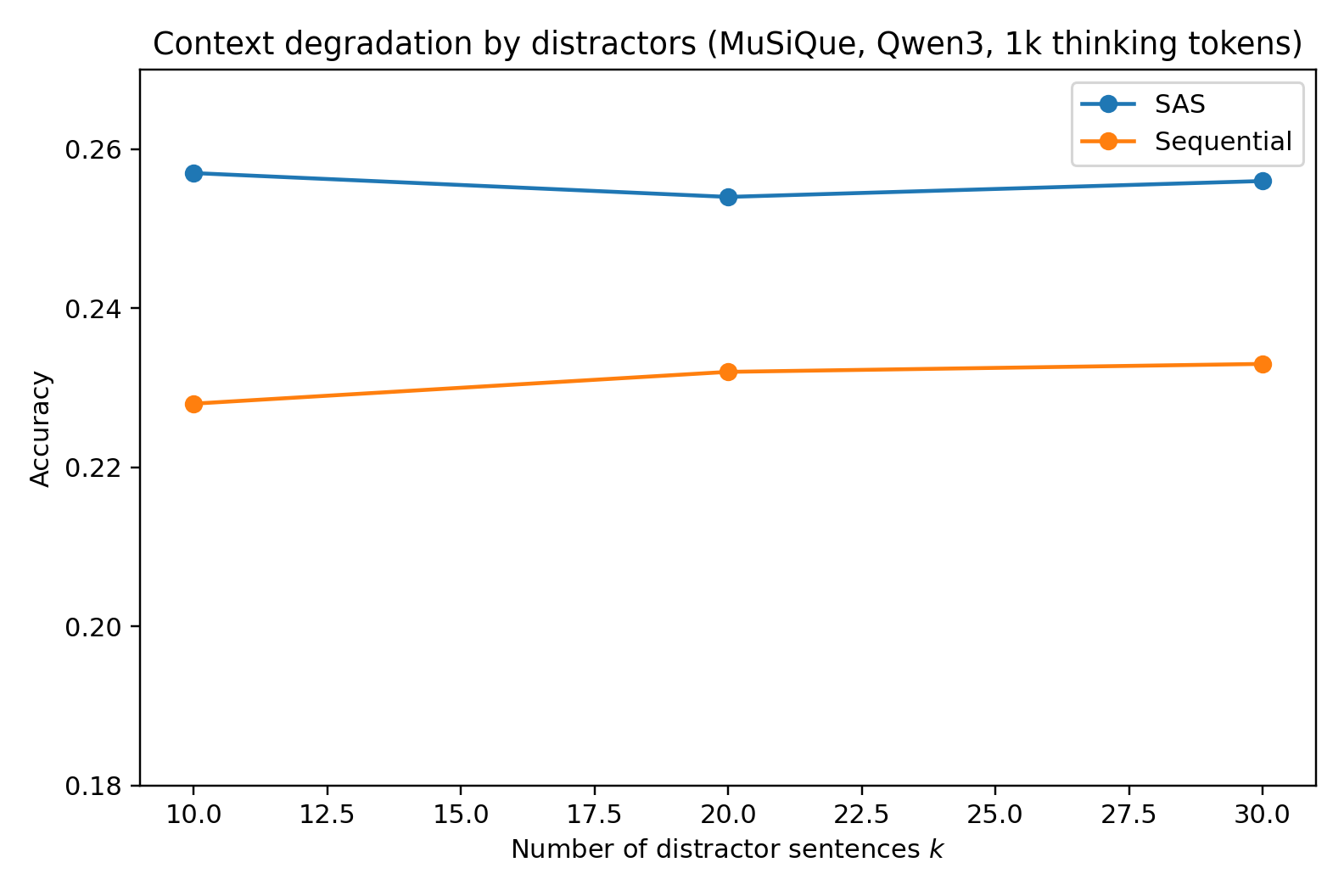}
    \end{minipage}
    \caption{Context degradation results on MuSiQue 4-hop with Qwen3-30B-A3B under a fixed 1000-token thinking budget. The x-axis is the degradation level ($\alpha$ for deletion, masking, and substitution; $k$ for distractors), and the y-axis is answer accuracy.}
    \label{fig:context-degradation-grid}
\end{figure*}

\subsection{Context degradation experiments}
\label{subsec:results-context-degradation}

Our theoretical analysis in Section~\ref{sec:theory_justification} predicts that multi-agent designs can become more competitive when the single agent’s effective access to the full context is degraded.
To probe this mechanism empirically, we explicitly control for \emph{context degradation} for Qwen3-30B-A3B on MuSiQue with a fixed 1000-token budget. 
For each degradation level we evaluate SAS and Sequential MAS, holding all other settings fixed.
Figure~\ref{fig:context-degradation-grid} visualizes the full pattern across all four degradation families.

We experiment with four alternative degradation approaches: randomly delete a fraction $\alpha$ of tokens from the context before inference (deletion); randomly replace a fraction $\alpha$ of tokens with a mask symbol (masking); randomly replace a fraction $\alpha$ of tokens with random vocabulary tokens (substitution); and append $k$ random distractor sentences that are topically similar but irrelevant to the question (distractor).

\textbf{The strongest evidence comes from degradations that corrupt information rather than merely remove it.} Under masking, SAS leads at mild degradation~($\alpha=0.3$) and the systems are roughly tied at moderate degradation, but Sequential becomes better at heavy degradation~($\alpha=0.7$). Under substitution, which both removes signal and injects misleading content, the crossover is even clearer: SAS is ahead at $\alpha=0.3$, the two systems move closer at $\alpha=0.5$, and Sequential is clearly better at $\alpha=0.7$. These are precisely the settings in which a structured multi-step pipeline is more robust to degradation through filtering, decomposing, or stabilizing reasoning that would otherwise be derailed by corrupted context.

Deletion produces a weaker version of the same trend. SAS remains stronger at mild degradation, the systems overlap at moderate degradation, and SAS regains a small edge at the heaviest deletion level. Adding distractor sentences is milder still: both systems degrade, but SAS remains ahead throughout, albeit by a smaller margin than in the clean setting. Taken together, these results suggest that MAS is most helpful not simply when context is longer, but when it becomes harder for a single reasoning trajectory to distinguish relevant from misleading information.

This degradation study complements the main results in an important way. Our claim is not that SAS always dominates in every regime. Rather, it is that \textbf{under matched budgets and proper context utilization, SAS is usually the strongest default}. The degradation experiments identify the boundary of that claim: \textbf{when effective single-agent context utilization deteriorates enough, structured multi-agent reasoning becomes competitive and can occasionally be better.} %This is precisely the regime where hybrid or context-aware agentic designs appear most promising.

\subsection{Additional experiments}

See Appendix \ref{appendix:error_analysis} for a detailed error analysis. The appendix also includes a detailed study of Gemini context utilization and a paraphrasing ablation study that tests the robustness of our findings to rewording and potential benchmark memorization, see Appendices \ref{appendix:gemini_accounting} and \ref{appendix:paraphrase} respectively.

\section{Conclusion}
We presented a budget-controlled comparison of single-agent~(SAS) and multi-agent~(MAS) LLM systems, focusing on fixed \emph{thinking token} budgets. Our results across two datasets~(FRAMES, MuSiQue), three model families~(Qwen3, DeepSeek, Gemini), and five different MAS architectures~(Sequential, Debate, Ensemble, Parallel-roles, Subtask-parallel) consistently show that SAS matches or outperforms MAS when computation is normalized, unless context utilization is degraded to a certain point. 

Overall, our results suggest that many reported MAS gains are better explained by compute and context effects than by inherent architectural superiority, and that future work should focus on the specific regimes where multi-agent structure provides real benefit.

\section{Acknowledgement}
We thank Dilip Arumugam for pointing out a small mistake in an earlier version of this work.

\bibliography{colm2026_conference}

@inproceedings{wang2024reasoning,
  title={Reasoning in token economies: budget-aware evaluation of LLM reasoning strategies},
  author={Wang, Junlin and Jain, Siddhartha and Zhang, Dejiao and Ray, Baishakhi and Kumar, Varun and Athiwaratkun, Ben},
  booktitle={Proceedings of the 2024 Conference on Empirical Methods in Natural Language Processing},
  pages={19916--19939},
  year={2024}
}

@inproceedings{han2025token,
  title={Token-budget-aware llm reasoning},
  author={Han, Tingxu and Wang, Zhenting and Fang, Chunrong and Zhao, Shiyu and Ma, Shiqing and Chen, Zhenyu},
  booktitle={Findings of the Association for Computational Linguistics: ACL 2025},
  pages={24842--24855},
  year={2025}
}

@article{li2025selfbudgeter,
  title={Selfbudgeter: Adaptive token allocation for efficient llm reasoning},
  author={Li, Zheng and Dong, Qingxiu and Ma, Jingyuan and Zhang, Di and Jia, Kai and Sui, Zhifang},
  journal={arXiv preprint arXiv:2505.11274},
  year={2025}
}

@article{wen2025budgetthinker,
  title={Budgetthinker: Empowering budget-aware llm reasoning with control tokens},
  author={Wen, Hao and Wu, Xinrui and Sun, Yi and Zhang, Feifei and Chen, Liye and Wang, Jie and Liu, Yunxin and Liu, Yunhao and Zhang, Ya-Qin and Li, Yuanchun},
  journal={arXiv preprint arXiv:2508.17196},
  year={2025}
}

@article{gao2025whynotboth,
  title={Single-agent or Multi-agent Systems? Why Not Both?},
  author={Gao, Mingyan and Li, Yanzi and Liu, Banruo and Yu, Yifan and Wang, Phillip and Lin, Ching-Yu and Lai, Fan},
  journal={arXiv preprint arXiv:2505.18286},
  year={2025}
}

@inproceedings{li2025single,
  title={Single-agent vs. multi-agent LLM strategies for automated student reflection assessment},
  author={Li, Gen and Chen, Li and Tang, Cheng and {\v{S}}v{\'a}bensk{\`y}, Valdemar and Deguchi, Daisuke and Yamashita, Takayoshi and Shimada, Atsushi},
  booktitle={Pacific-Asia Conference on Knowledge Discovery and Data Mining},
  pages={300--311},
  year={2025},
  organization={Springer}
}

@article{zhu2025medagentboard,
  title={Medagentboard: Benchmarking multi-agent collaboration with conventional methods for diverse medical tasks},
  author={Zhu, Yinghao and He, Ziyi and Hu, Haoran and Zheng, Xiaochen and Zhang, Xichen and Wang, Zixiang and Gao, Junyi and Ma, Liantao and Yu, Lequan},
  journal={arXiv preprint arXiv:2505.12371},
  year={2025}
}

@inproceedings{krishna2025fact,
  title={Fact, fetch, and reason: A unified evaluation of retrieval-augmented generation},
  author={Krishna, Satyapriya and Krishna, Kalpesh and Mohananey, Anhad and Schwarcz, Steven and Stambler, Adam and Upadhyay, Shyam and Faruqui, Manaal},
  booktitle={Proceedings of the 2025 Conference of the Nations of the Americas Chapter of the Association for Computational Linguistics: Human Language Technologies (Volume 1: Long Papers)},
  pages={4745--4759},
  year={2025}
}

@article{trivedi2022musique,
  title={MuSiQue: Multihop Questions via Single-hop Question Composition},
  author={Trivedi, Harsh and Balasubramanian, Niranjan and Khot, Tushar and Sabharwal, Ashish},
  journal={Transactions of the Association for Computational Linguistics},
  volume={10},
  pages={539--554},
  year={2022},
  publisher={MIT Press One Broadway, 12th Floor, Cambridge, Massachusetts 02142, USA~…}
}

@book{CoverThomas2006,
  title={Elements of Information Theory},
  author={Cover, Thomas M. and Thomas, Joy A.},
  year={2006},
  publisher={Wiley-Interscience},
  edition={2nd}
}

@article{du2025context,
  title={Context length alone hurts LLM performance despite perfect retrieval},
  author={Du, Yufeng and Tian, Minyang and Ronanki, Srikanth and Rongali, Subendhu and Bodapati, Sravan and Galstyan, Aram and Wells, Azton and Schwartz, Roy and Huerta, Eliu A and Peng, Hao},
  journal={arXiv preprint arXiv:2510.05381},
  year={2025}
}

@inproceedings{li2025focusllm,
  title={FocusLLM: Precise understanding of long context by dynamic condensing},
  author={Li, Zhenyu and Zhang, Yike and Pan, Tengyu and Sun, Yutao and Duan, Zhichao and Fang, Junjie and Han, Rong and Wang, Zixuan and Wang, Jianyong},
  booktitle={Proceedings of the 63rd Annual Meeting of the Association for Computational Linguistics (Volume 1: Long Papers)},
  pages={31087--31101},
  year={2025}
}

@article{lu2024controlled,
  title={A controlled study on long context extension and generalization in llms},
  author={Lu, Yi and Yan, Jing Nathan and Yang, Songlin and Chiu, Justin T and Ren, Siyu and Yuan, Fei and Zhao, Wenting and Wu, Zhiyong and Rush, Alexander M},
  journal={arXiv preprint arXiv:2409.12181},
  year={2024}
}

@article{li2024more,
  title={More agents is all you need},
  author={Li, Junyou and Zhang, Qin and Yu, Yangbin and Fu, Qiang and Ye, Deheng},
  journal={arXiv preprint arXiv:2402.05120},
  year={2024}
}

@inproceedings{du2024improving,
  title={Improving factuality and reasoning in language models through multiagent debate},
  author={Du, Yilun and Li, Shuang and Torralba, Antonio and Tenenbaum, Joshua B and Mordatch, Igor},
  booktitle={Forty-first international conference on machine learning},
  year={2024}
}

@article{zhang2023exploring,
  title={Exploring collaboration mechanisms for llm agents: A social psychology view},
  author={Zhang, Jintian and Xu, Xin and Zhang, Ningyu and Liu, Ruibo and Hooi, Bryan and Deng, Shumin},
  journal={arXiv preprint arXiv:2310.02124},
  year={2023}
}

@article{shinn2023reflexion,
  title={Reflexion: Language agents with verbal reinforcement learning},
  author={Shinn, Noah and Cassano, Federico and Gopinath, Ashwin and Narasimhan, Karthik and Yao, Shunyu},
  journal={Advances in neural information processing systems},
  volume={36},
  pages={8634--8652},
  year={2023}
}

@article{kim2025towards,
  title={Towards a science of scaling agent systems},
  author={Kim, Yubin and Gu, Ken and Park, Chanwoo and Park, Chunjong and Schmidgall, Samuel and Heydari, A Ali and Yan, Yao and Zhang, Zhihan and Zhuang, Yuchen and Malhotra, Mark and others},
  journal={arXiv preprint arXiv:2512.08296},
  year={2025}
}

@article{cemri2025multi,
  title={Why do multi-agent llm systems fail?},
  author={Cemri, Mert and Pan, Melissa Z and Yang, Shuyi and Agrawal, Lakshya A and Chopra, Bhavya and Tiwari, Rishabh and Keutzer, Kurt and Parameswaran, Aditya and Klein, Dan and Ramchandran, Kannan and others},
  journal={arXiv preprint arXiv:2503.13657},
  year={2025}
}

@article{ke2026mas,
  title={Mas-orchestra: Understanding and improving multi-agent reasoning through holistic orchestration and controlled benchmarks},
  author={Ke, Zixuan and Ming, Yifei and Xu, Austin and Chin, Ryan and Nguyen, Xuan-Phi and Jwalapuram, Prathyusha and Wang, Jiayu and Yavuz, Semih and Xiong, Caiming and Joty, Shafiq},
  journal={arXiv preprint arXiv:2601.14652},
  year={2026}
}

@article{liu2024lost,
  title={Lost in the middle: How language models use long contexts},
  author={Liu, Nelson F and Lin, Kevin and Hewitt, John and Paranjape, Ashwin and Bevilacqua, Michele and Petroni, Fabio and Liang, Percy},
  journal={Transactions of the association for computational linguistics},
  volume={12},
  pages={157--173},
  year={2024}
}

@techreport{hong2025contextrot,
  title={Context Rot: How Increasing Input Tokens Impacts LLM Performance},
  author={Hong, Kelly and Troynikov, Anton and Huber, Jeff},
  institution={Chroma},
  year={2025},
  month={July},
  url={https://www.trychroma.com/research/context-rot}
}

@misc{anthropic2025researchsystem,
  author={{Anthropic}},
  title={How We Built Our Multi-Agent Research System},
  year={2025},
  month=jun,
  day={13},
  howpublished = {\url{https://www.anthropic.com/engineering/multi-agent-research-system}}
}
\bibliographystyle{colm2026_conference}

\appendix

\section{Paraphrasing Ablation Study}
\label{appendix:paraphrase}
To investigate the impact of potential dataset contamination or question phrasing artifacts, we ran an ablation on the MuSiQue 4-hop dataset using two paraphrasing methods on the questions.

\paragraph{Paraphrasing Methods:}
We implemented two distinct methods:
\begin{itemize}
    \item \textbf{Light Paraphrase:} A rule-based approach that uses regular expressions to swap common question phrases (e.g., "When was" $\to$ "At what time was", "founded" $\to$ "established") without altering the core structure.
    \item \textbf{Deep Paraphrase:} An LLM-based approach using Gemini-2.5-Flash. The model was given a system prompt with hard rules to completely rephrase the question while preserving the exact meaning, answer, and multi-hop structure, aiming for low lexical overlap with the original.
\end{itemize}

We evaluated SAS and Sequential MAS configurations on these two new datasets at 1k and 2k thinking token budgets. Table~\ref{tab:appendix-paraphrase} summarizes all the representative values.

\paragraph{Discussion of Paraphrase Results:}
The results show two interesting trends across both models. First, the \textbf{Light Paraphrase} consistently \emph{decreased} performance for SAS (For Gemini: .331$\to$.326; For Qwen3: .260$\to$.249). This was particularly notable for Sequential MAS in Qwen3, which saw a significant drop (.229$\to$.204). This suggests that the simple regex-based changes may have obfuscated the questions or removed helpful phrasing cues that the models were relying on.

Second, the \textbf{Deep Paraphrase} \emph{improved} performance, especially for the stronger Gemini model. For Gemini-2.5-Flash, SAS performance jumped from .331 to .358 at 1k tokens. Qwen3 also saw slight performance gains in both SAS and Sequential MAS, even though it is not significant. This divergence, where simple paraphrasing hurts and deep paraphrasing helps, strongly suggests that the original questions may suffer from memorization or overfitting from pretraining. The deep, semantically-equivalent rephrasing forces the model to perform more robust reasoning, supporting the use of this technique for creating more reliable agent benchmarks.

\begin{table*}[t]
\centering
\caption{Error analysis on MuSiQue 4-hop at 1k thinking tokens, comparing Gemini-2.5-Flash and Qwen3-30B-A3B. The MAS being used in here is Sequential MAS. Buckets are: MAS right \& SAS wrong (MR/SW), SAS right \& MAS wrong (SR/MW), both right (BR), and both wrong (BW). ``Avg Tokens'' gives the average thinking-token count; ``Entity Spans'' is the average number of entities traversed in the internal reasoning; ``\% gold in thoughts'' is the percentage of examples in which the gold answer appears anywhere in the internal reasoning.}
\label{tab:error-analysis-merged}
\resizebox{\linewidth}{!}{%
\begin{tabular}{l|rrrrrr|rrrrrr}
\toprule
& \multicolumn{6}{c|}{\textbf{Gemini-2.5-Flash}} & \multicolumn{6}{c}{\textbf{Qwen3-30B-A3B}} \\
\cmidrule(lr){2-7}\cmidrule(lr){8-13}
\textbf{Bucket}
& $N$
& \multicolumn{2}{c}{Avg Tokens}
& \multicolumn{2}{c}{Entity Spans}
& \% gold in thoughts
& $N$
& \multicolumn{2}{c}{Avg Tokens}
& \multicolumn{2}{c}{Entity Spans}
& \% gold in thoughts \\
\cmidrule(lr){3-4}\cmidrule(lr){5-6}\cmidrule(lr){9-10}\cmidrule(lr){11-12}
& & SAS & Seq & SAS & Seq & SAS / Seq / Both
& & SAS & Seq & SAS & Seq & SAS / Seq / Both \\
\midrule
MR/SW
& 72  & 260 & 516 & 4.8 & 9.5 & 12.5 / 41.7 / ---
& 60  & 788 & 769 & 9.3 & 9.9 & 18.3 / 56.7 / --- \\
SR/MW
& 124 & 286 & 565 & 5.9 & 10.3 & 42.7 / 18.6 / ---
& 96  & 733 & 791 & 8.1 & 9.2 & 63.5 / 28.1 / --- \\
BR
& 265 & 217 & 483 & 5.0 & 10.0 & 61.9 / 64.5 / 50.9
& 209 & 666 & 761 & 7.8 & 9.4 & 73.7 / 78.5 / 69.9 \\
BW
& 714 & 274 & 542 & 5.2 & 10.1 & 3.36 / 2.66 / ---
& 810 & 787 & 809 & 7.8 & 8.9 & 3.33 / 3.58 / --- \\
\bottomrule
\end{tabular}
}
\end{table*}

\section{Error Analysis}
\label{appendix:error_analysis}

In order to understand our results better, we performed a fine-grained analysis of \emph{unfiltered thinking} for Gemini-2.5-Flash and Qwen3-30B-A3B at  1k thinking tokens on MuSiQue 4-hop questions, partitioning examples into four mutually exclusive buckets: (\textbf{MR/SW}) Sequential MAS right \& SAS wrong, (\textbf{SR/MW}) SAS right \& Sequential MAS wrong, (\textbf{BR}) Both right, and (\textbf{BW}) Both wrong. We summarize corpus-level properties of the internal chains and identify error patterns that consistently predict success or failure in Table \ref{tab:error-analysis-merged}. We refer to the ground truth answer here as "gold".

\paragraph{MR/SW (Sequential MAS right, SAS wrong).}
Sequential MAS writes longer thoughts and canvasses 2 times more entities than SAS does; the correct gold appears in Sequential MAS’s thoughts far more often, compared with SAS (41.7\% vs.\ 12.5\% for Gemini, and 56.7\% vs. 18.3\% for Qwen3). We also observe \textit{extraction failures} on the SAS side: the gold appears in SAS’s thoughts but is dropped in the final answer in 9 out of 72 cases in Gemini case.  
\textbf{Interpretation:} SAS \emph{under-explores} and sometimes fails to copy the correct span from its own chain; Sequential MAS’s breadth plus occasional backtracking rescues the final.

\paragraph{SR/MW (SAS right, Sequential MAS wrong).}
SAS’s chains maintain a much higher lexical overlap with the question; the gold string appears in SAS’s thoughts in 42.7\% vs.\ 18.6\% for MAS for Gemini, and 63.5\% vs.\ 28.1\% for Qwen 3. Sequential MAS \emph{over-explores} and \emph{drifts}, including 23 cases of \textit{extraction failure} for Gemini (gold surfaced in thoughts but not in the final).  
\textbf{Interpretation:} Tighter \emph{constraint anchoring} in SAS is decisive. Sequential MAS’s breadth, absent pruning, degrades precision and sometimes loses a correct span at finalization.

\paragraph{BR (Both right).}
Both systems frequently contain the gold in thoughts (SAS 61.9\%, Sequential MAS 64.5\%, both 50.9\% for Gemini; SAS 73.7\%, Sequential MAS 78.5\%, both 69.9\% for Qwen3). Backtracking markers occur on both sides, suggesting explicit course correction before convergence. 
\textbf{Interpretation:} Success typically follows a two-stage path: (i) surface a plausible span, (ii) perform a late constraint re-check that avoids switching away from the correct span.

\paragraph{BW (Both wrong).}
The gold string almost never appears (single 3.36\%, multi 2.66\% for Gemini; single 3.33\%, multi 3.58\% for Qwen3). Both systems pursue \emph{disjoint} candidates and fail to reconcile.  Both produce long, but misguided thoughts. 
\textbf{Interpretation:} The dominant failure mode is \textit{mutual drift} with insufficient late binding to the question’s constraints; neither path surfaces an extractable gold span.

\paragraph{Model-specific notes.}
Gemini-2.5-Flash generally allocates substantially more thinking text to Sequential MAS than SAS across buckets, magnifying the breadth vs.\ precision pattern.
Qwen3-30B-A3B shows a similar picture, with Sequential MAS exploring more entities than SAS across buckets; the relative token lengths are closer in some buckets, but the \emph{gold-in-thoughts percentage} remains the main predictor of whether one can answer the question correctly or not.

\paragraph{Takeaway.}
With this analysis, we can see that SAS succeeds by \emph{staying close to the question} and reliably carrying the surfaced span into the final, while Sequential MAS succeeds when its extra breadth is paired with late constraint checking. Failures for both are dominated by: (i) never writing the correct surface form, and (ii) losing a previously correct span at finalization.

\section{Limitations}
\label{sec:limitations}
(i) We focus on text-only multi-hop reasoning; MAS advantages with tools/vision or safety constraints are out of scope. (ii) Gemini thinking accounting is approximate; we report both API and content-based proxies and emphasize accuracy under matched \emph{requested} budgets. (iii) We only study the effect on performance across model families and datasets while increasing the thinking token cap. We do not enforce the models to actually use up all of those budgets. As a result, we restrict only upper thinking budget cap, not lower thinking budget cap.

\section{Architecture Prompts}
\label{appendix:architecture-prompts}

This section lists all the prompts used in the paper. Each subsection corresponds to one architecture and groups its prompts by function. Except for temperature, all other hyperparameters are kept default. Temperature value used for Ensemble is 0.7, otherwise 0 for the rest.

\subsection{Single-Agent System}
This subsection contains the prompts used for the standard single-agent system and the longer-thinking single-agent variant.

\subsubsection{SAS system prompt}
\begin{lstlisting}[style=promptstyle]
You are a helpful assistant. Think step by step, then answer.
Be as succinct as possible. Do NOT repeat the question.
Return ONLY the final answer requested.
\end{lstlisting}

\subsubsection{SAS-L added user prefix}
\begin{lstlisting}[style=promptstyle]
I want you to analyze the following question from multiple perspectives before answering.

1. Identify ambiguities.
2. Formulate at least two plausible interpretations.
3. Evaluate the interpretations and choose the most likely one.
4. Answer based on the most likely interpretation.

The question is:
\end{lstlisting}

\subsection{Sequential Multi-Agent System}
This subsection contains the prompts used for the Sequential planner, step agents, and aggregator.

\subsubsection{Planner system prompt}
\begin{lstlisting}[style=promptstyle]
You are a careful planner. Break the user task into the fewest necessary sequential steps so each step output feeds the next.

Output strict JSON only with the following structure:
steps:
  - id: 1
    name: Step 1
    instruction: What to do
  - id: 2
    name: Step 2
    instruction: What to do

Rules:
- Steps must be sequential and minimal.
- Do not answer the question yourself.
- Instructions must be concrete and unambiguous.
- Do not include commentary outside the JSON object.
\end{lstlisting}

\subsubsection{Per-step agent system prompt}
\begin{lstlisting}[style=promptstyle]
You are a helpful assistant. Think only for your step.
\end{lstlisting}

\subsubsection{Per-step agent user template}
\begin{lstlisting}[style=promptstyle]
Original Question: {q}

Full Plan:
{plan_as_text}

You are responsible for Step {i}: {step_name}
Instruction: {step_instruction}

Previous step outputs:
{prior_step_outputs}

Perform ONLY your assigned step. Provide your step output succinctly.
\end{lstlisting}

\subsubsection{Aggregator system prompt}
\begin{lstlisting}[style=promptstyle]
You are an aggregator. Read step outputs and return the final answer only.
Do NOT attempt to solve the question yourself.
\end{lstlisting}

\subsubsection{Aggregator user template}
\begin{lstlisting}[style=promptstyle]
Question: {q}

Step outputs:
{step_outputs}

Return the final answer only.
\end{lstlisting}

\subsection{Subtask-Parallel Multi-Agent System}
This subsection contains the prompts used for subtask decomposition, parallel workers, and aggregator.

\subsubsection{Planner system prompt}
\begin{lstlisting}[style=promptstyle]
You are a planner. Decompose the question into a small set of independent subtasks that can be solved in parallel.

Output strict JSON only with the following structure:
tasks:
  - id: 1
    name: Task 1
    instruction: What to do
    deliverable: What to return

Rules:
- Tasks must be independent.
- Keep tasks minimal and directly useful.
- Do not answer the question yourself.
- Do not include commentary outside the JSON object.
\end{lstlisting}

\subsubsection{Worker system prompt}
\begin{lstlisting}[style=promptstyle]
You are a helpful assistant.
\end{lstlisting}

\subsubsection{Worker user template}
\begin{lstlisting}[style=promptstyle]
Question: {q}

Task {task_id}:
Instruction: {task_instruction}
Deliverable: {task_deliverable}

Return only what the deliverable asks for.
\end{lstlisting}

\subsubsection{Aggregator system prompt}
\begin{lstlisting}[style=promptstyle]
You are a reducer. You will be given outputs from multiple subtasks.
Synthesize them into the best possible final answer.
Return only the final answer.
\end{lstlisting}

\subsubsection{Aggregator user template}
\begin{lstlisting}[style=promptstyle]
Question: {q}

Task outputs:
{task_outputs}

Return the final answer only.
\end{lstlisting}

\subsection{Parallel-Roles Multi-Agent System}
This subsection contains the prompts used for role-specialized workers and aggregator.

\subsubsection{Worker system prompt}
\begin{lstlisting}[style=promptstyle]
You are a role-specialized assistant.
Follow the assigned role instructions carefully and produce the best partial answer for that role.
\end{lstlisting}

\subsubsection{Worker user template}
\begin{lstlisting}[style=promptstyle]
Question: {q}

Role: {role_name}

Role instructions:
{instructions}

Produce your analysis and the best partial answer for this role.
\end{lstlisting}

\subsubsection{Aggregator system prompt}
\begin{lstlisting}[style=promptstyle]
You are an aggregator. Read the worker outputs and return the final answer only.
Do not add commentary.
\end{lstlisting}

\subsubsection{Aggregator user template}
\begin{lstlisting}[style=promptstyle]
Question: {q}

Role outputs:
{role_outputs}

Return only the final answer.
\end{lstlisting}

\subsection{Debate Multi-Agent System}
This subsection contains the prompts used for the debaters, critics, and aggregator.

\subsubsection{Debater system prompt}
\begin{lstlisting}[style=promptstyle]
You are a debater. Provide the best possible answer to the question.
Think step by step in private, then output only the final answer.
\end{lstlisting}

\subsubsection{Critic system prompt}
\begin{lstlisting}[style=promptstyle]
You are a critic. You will be given an opponent answer.
Point out flaws, missing constraints, or alternative reasoning.
Then provide a corrected improved answer.
Output only the final improved answer.
\end{lstlisting}

\subsubsection{Critique user template}
\begin{lstlisting}[style=promptstyle]
Opponent answer:
{opponent_answer}

Your critique and improved final answer:
\end{lstlisting}

\subsubsection{Aggregator system prompt}
\begin{lstlisting}[style=promptstyle]
You are a judge. You will be given two candidate answers.
Select the better one. If both are wrong, pick the one that is closer.
Output only the final answer.
\end{lstlisting}

\subsubsection{Aggregator user template}
\begin{lstlisting}[style=promptstyle]
Question: {q}

Answer A:
{answer_a}

Answer B:
{answer_b}

Pick the better answer and output only the final answer.
\end{lstlisting}

\subsection{Ensemble Multi-Agent System}
This subsection contains the prompts used for independent candidates and aggregator.

\subsubsection{Candidate worker system prompt}
\begin{lstlisting}[style=promptstyle]
You are a helpful assistant. Solve the question independently and return only the final answer.
\end{lstlisting}

\subsubsection{Aggregator system prompt}
\begin{lstlisting}[style=promptstyle]
You are a judge. You will be given a question and multiple candidate answers.
Pick the single best answer. If all are wrong, pick the closest.
Output only the final answer.
\end{lstlisting}

\subsubsection{Aggregator user template}
\begin{lstlisting}[style=promptstyle]
Question: {q}

Candidates:
{candidate_answers}

Output only the final answer.
\end{lstlisting}

\subsection{Evaluation}
This subsection contains the prompts used for evaluation. The evaluation step is ran right after the predicted output is produced, using the same model.

\subsubsection{Evaluation System prompt}
\begin{lstlisting}[style=promptstyle]
You are a helpful assistant.
\end{lstlisting}

\subsubsection{Evaluation User prompt}
\begin{lstlisting}[style=promptstyle]
===Task===

I need your help in evaluating an answer provided by an LLM against a ground truth answer.
Your task is to determine if the ground truth answer is present in the LLM's response.
Please analyze the provided data and make a decision.

===Instructions===
1. Carefully compare the Predicted Answer with the Ground Truth Answer.
2. The Ground Truth Answer is always absolutely correct. Do NOT assume otherwise.
3. Consider the substance of the answers - look for equivalent information or correct answers.
   Do not focus on exact wording unless the exact wording is crucial to the meaning.
4. Your final decision should be based on whether the meaning and the vital facts of the
   Ground Truth Answer are present in the Predicted Answer:

===Input Data===
- Question: <question>
- Predicted Answer: <LLM_response>
- Ground Truth Answer: <ground_truth_answer>

===Output Format===
Provide your final evaluation in the following dictionary format:
{Explanation: <How you made the decision?>, Decision: <TRUE or FALSE>}
\end{lstlisting}

\section{Context Degradation Logic}
\label{appendix:context-degradation}

This section documents the context degradation experiments. All four methods use the same generation workflow and differ only in how the generated thought text is corrupted before answer generation.

\subsection{Common Workflow}
This subsection describes the common generation logic shared by deletion, masking, substitution, and distractor insertion.

\begin{lstlisting}[style=promptstyle]
Input:
- messages
- thinking budget B
- corruption method T

Step 1:
Build the chat prelude and append the opening think tag.

Step 2:
Generate think text with a hard cap of B thinking tokens.

Step 3:
Apply corruption operator T only to the generated think text.

Step 4:
Reconstruct the prompt as:
  prelude
  corrupted think text
  closing think tag

Step 5:
Generate the final answer from the corrupted reasoning context.

Output:
- corrupted full trace
- used thinking tokens
- corruption metadata
\end{lstlisting}

\subsection{Deletion}
This subsection describes the deletion corruption logic used in the deletion experiments.

\begin{lstlisting}[style=promptstyle]
Parameters:
- delete_pct
- every_n
- seed

Input:
- clean think text

If corruption is disabled:
  return clean think text

Else:
  initialize RNG with base seed plus call index

  If every_n is set:
    delete every n-th word deterministically
  Else:
    delete approximately delete_pct of words at random

Return:
- corrupted think text
- metadata containing method, seed, and parameters
\end{lstlisting}

\subsection{Masking}
This subsection describes the masking corruption logic used in the masking experiments.

\begin{lstlisting}[style=promptstyle]
Parameters:
- mask_pct
- mask_token
- every_n
- seed

Input:
- clean think text

If corruption is disabled:
  return clean think text

Else:
  initialize RNG with base seed plus call index

  If every_n is set:
    replace every n-th word with the mask token
  Else:
    replace approximately mask_pct of words with the mask token

Return:
- corrupted think text
- metadata containing method, seed, and parameters
\end{lstlisting}

\subsection{Substitution}
This subsection describes the substitution corruption logic used in the substitution experiments.

\begin{lstlisting}[style=promptstyle]
Parameters:
- replace_pct
- vocab_sample_size
- avoid_special
- seed

Input:
- clean think text
- tokenizer

If corruption is disabled:
  return clean think text

Else:
  initialize RNG with base seed plus call index
  sample replacement tokens from the tokenizer vocabulary
  optionally exclude special tokens
  replace approximately replace_pct of tokens in the think text

Return:
- corrupted think text
- metadata containing method, seed, and parameters
\end{lstlisting}

\subsection{Distractor Insertion}
This subsection describes the distractor insertion logic used in the distractor experiments.

\begin{lstlisting}[style=promptstyle]
Parameters:
- num_distractors
- distractor pool
- seed

Input:
- clean think text

If corruption is disabled:
  return clean think text

Else:
  initialize RNG with base seed plus call index
  sample num_distractors distractor sentences
  insert distractors into the think text

Return:
- corrupted think text
- metadata containing method, seed, and parameters
\end{lstlisting}

\section{Full results with bootstrap confidence intervals}
\label{appendix:confidence}
This section reports the full numerical results corresponding to the experiments summarized in the main paper. We include all reported accuracies together with their 95\% bootstrap confidence intervals.

\subsection{Open-source models}
This subsection reports the open-source models' main results together with average thinking token usage and 95\% bootstrap confidence intervals (Table \ref{tab:appendix-qwen-frames}, Table \ref{tab:appendix-qwen-musique}, Table \ref{tab:appendix-deepseek-frames}, and Table \ref{tab:appendix-deepseek-musique}).

\begin{table*}[t]
\centering
\caption{FRAMES full results for Qwen3-30B-A3B.}
\label{tab:appendix-qwen-frames}
\resizebox{\linewidth}{!}{%
\begin{tabular}{lcccccc}
\toprule
System & 100 & 500 & 1k & 2k & 5k & 10k \\
\midrule
SAS & 0.191 & 0.240 & 0.252 & 0.250 & 0.260 & 0.263 \\
95\% CI & (0.158, 0.201) & (0.238, 0.268) & (0.251, 0.282) & (0.248, 0.275) & (0.247, 0.295) & (0.252, 0.299) \\
Avg. thinking tokens & 100 & 474 & 800 & 1103 & 1260 & 1307 \\
Seq & 0.198 & 0.223 & 0.225 & 0.252 & 0.252 & 0.258 \\
95\% CI & (0.174, 0.206) & (0.214, 0.224) & (0.205, 0.232) & (0.227, 0.255) & (0.229, 0.253) & (0.247, 0.281) \\
Avg. thinking tokens & 99.5 & 494 & 889 & 1321 & 1693 & 1808 \\
SAS-L & 0.195 & 0.220 & 0.235 & 0.246 & 0.246 & 0.246 \\
95\% CI & (0.172, 0.210) & (0.204, 0.250) & (0.223, 0.253) & (0.233, 0.261) & (0.237, 0.261) & (0.237, 0.261) \\
Avg. thinking tokens & 100 & 500 & 935 & 1232 & 1309 & 1327 \\
Debate & 0.204 & 0.206 & 0.228 & 0.232 & 0.238 & 0.240 \\
95\% CI & (0.178, 0.209) & (0.173, 0.217) & (0.202, 0.236) & (0.202, 0.241) & (0.210, 0.253) & (0.217, 0.242) \\
Avg. thinking tokens & 100 & 500 & 1000 & 1944 & 3414 & 4061 \\
Ensemble & 0.146 & 0.193 & 0.210 & 0.230 & 0.254 & 0.263 \\
95\% CI & (0.117, 0.149) & (0.177, 0.195) & (0.184, 0.211) & (0.206, 0.253) & (0.215, 0.262) & (0.221, 0.277) \\
Avg. thinking tokens & 100 & 500 & 1000 & 1964 & 4009 & 5571 \\
Subtask-parallel & 0.155 & 0.187 & 0.220 & 0.232 & 0.237 & 0.244 \\
95\% CI & (0.136, 0.161) & (0.172, 0.197) & (0.205, 0.237) & (0.214, 0.233) & (0.225, 0.239) & (0.217, 0.251) \\
Avg. thinking tokens & 100 & 499 & 969 & 1670 & 2592 & 3001 \\
Parallel-roles & 0.207 & 0.223 & 0.240 & 0.256 & 0.265 & 0.271 \\
95\% CI & (0.168, 0.215) & (0.181, 0.229) & (0.213, 0.240) & (0.233, 0.261) & (0.245, 0.274) & (0.250, 0.284) \\
Avg. thinking tokens & 100 & 500 & 999 & 1934 & 3854 & 4850 \\
\bottomrule
\end{tabular}}
\end{table*}

\begin{table*}[t]
\centering
\caption{MuSiQue 4-hop full results for Qwen3-30B-A3B.}
\label{tab:appendix-qwen-musique}
\resizebox{\linewidth}{!}{%
\begin{tabular}{lcccccc}
\toprule
System & 100 & 500 & 1k & 2k & 5k & 10k \\
\midrule
SAS & 0.200 & 0.250 & 0.260 & 0.271 & 0.271 & 0.271 \\
95\% CI & (0.196, 0.214) & (0.248, 0.255) & (0.260, 0.271) & (0.269, 0.276) & (0.268, 0.276) & (0.268, 0.276) \\
Avg. thinking tokens & 100 & 495 & 916 & 1345 & 1453 & 1462 \\
Seq & 0.220 & 0.226 & 0.229 & 0.229 & 0.231 & 0.231 \\
95\% CI & (0.212, 0.238) & (0.229, 0.235) & (0.223, 0.248) & (0.223, 0.240) & (0.226, 0.243) & (0.225, 0.244) \\
Avg. thinking tokens & 99.6 & 495 & 907 & 1383 & 1766 & 1811 \\
SAS-L & 0.210 & 0.213 & 0.231 & 0.239 & 0.248 & 0.248 \\
95\% CI & (0.200, 0.214) & (0.211, 0.224) & (0.227, 0.238) & (0.238, 0.255) & (0.248, 0.271) & (0.248, 0.271) \\
Avg. thinking tokens & 100 & 500 & 976 & 1340 & 1379 & 1379 \\
Debate & 0.225 & 0.229 & 0.224 & 0.234 & 0.249 & 0.244 \\
95\% CI & (0.211, 0.244) & (0.216, 0.234) & (0.218, 0.228) & (0.225, 0.240) & (0.233, 0.255) & (0.237, 0.250) \\
Avg. thinking tokens & 100 & 500 & 1000 & 1968 & 3702 & 4465 \\
Ensemble & 0.149 & 0.183 & 0.197 & 0.224 & 0.226 & 0.245 \\
95\% CI & (0.137, 0.157) & (0.175, 0.204) & (0.194, 0.208) & (0.212, 0.229) & (0.209, 0.232) & (0.237, 0.254) \\
Avg. thinking tokens & 100 & 500 & 1000 & 1992 & 4523 & 6230 \\
Subtask-parallel & 0.174 & 0.187 & 0.207 & 0.234 & 0.249 & 0.254 \\
95\% CI & (0.156, 0.175) & (0.170, 0.196) & (0.197, 0.214) & (0.220, 0.237) & (0.243, 0.259) & (0.251, 0.261) \\
Avg. thinking tokens & 100 & 499 & 986 & 1882 & 3092 & 3510 \\
Parallel-roles & 0.202 & 0.204 & 0.220 & 0.226 & 0.246 & 0.242 \\
95\% CI & (0.193, 0.209) & (0.192, 0.217) & (0.214, 0.231) & (0.206, 0.241) & (0.233, 0.249) & (0.233, 0.251) \\
Avg. thinking tokens & 100 & 500 & 1000 & 1986 & 4420 & 5785 \\
\bottomrule
\end{tabular}}
\end{table*}

\begin{table*}[t]
\centering
\caption{FRAMES full results for DeepSeek-R1-Distill-Llama-70B.}
\label{tab:appendix-deepseek-frames}
\resizebox{\linewidth}{!}{%
\begin{tabular}{lcccccc}
\toprule
System & 100 & 500 & 1k & 2k & 5k & 10k \\
\midrule
SAS & 0.365 & 0.427 & 0.448 & 0.454 & 0.455 & 0.456 \\
95\% CI & (0.359, 0.378) & (0.407, 0.432) & (0.421, 0.459) & (0.431, 0.463) & (0.434, 0.464) & (0.434, 0.465) \\
Avg. thinking tokens & 100 & 466 & 550 & 574 & 635 & 960 \\
Seq & 0.387 & 0.396 & 0.391 & 0.393 & 0.394 & 0.397 \\
95\% CI & (0.365, 0.395) & (0.373, 0.397) & (0.360, 0.400) & (0.369, 0.405) & (0.372, 0.399) & (0.374, 0.400) \\
Avg. thinking tokens & 99.6 & 493 & 749 & 959 & 1080 & 1337 \\
SAS-L & 0.374 & 0.432 & 0.448 & 0.451 & 0.444 & 0.445 \\
95\% CI & (0.370, 0.385) & (0.423, 0.441) & (0.429, 0.452) & (0.439, 0.457) & (0.428, 0.450) & (0.414, 0.450) \\
Avg. thinking tokens & 100 & 459 & 567 & 698 & 755 & 860 \\
Debate & 0.400 & 0.392 & 0.397 & 0.425 & 0.439 & 0.444 \\
95\% CI & (0.359, 0.419) & (0.363, 0.406) & (0.374, 0.409) & (0.387, 0.432) & (0.414, 0.442) & (0.417, 0.448) \\
Avg. thinking tokens & 100 & 500 & 1000 & 1805 & 2376 & 2496 \\
Ensemble & 0.365 & 0.400 & 0.419 & 0.431 & 0.450 & 0.458 \\
95\% CI & (0.339, 0.395) & (0.384, 0.405) & (0.376, 0.426) & (0.402, 0.438) & (0.423, 0.457) & (0.442, 0.455) \\
Avg. thinking tokens & 100 & 500 & 999 & 1868 & 2879 & 3163 \\
Subtask-parallel & 0.385 & 0.402 & 0.420 & 0.434 & 0.436 & 0.434 \\
95\% CI & (0.369, 0.397) & (0.371, 0.419) & (0.393, 0.430) & (0.402, 0.443) & (0.400, 0.443) & (0.400, 0.441) \\
Avg. thinking tokens & 100 & 492 & 884 & 1261 & 1484 & 1559 \\
Parallel-roles & 0.427 & 0.407 & 0.425 & 0.433 & 0.448 & 0.450 \\
95\% CI & (0.411, 0.441) & (0.380, 0.409) & (0.397, 0.433) & (0.408, 0.434) & (0.433, 0.461) & (0.436, 0.465) \\
Avg. thinking tokens & 100 & 500 & 996 & 1814 & 2647 & 2627 \\
\bottomrule
\end{tabular}}
\end{table*}

\begin{table*}[t]
\centering
\caption{MuSiQue 4-hop full results for DeepSeek-R1-Distill-Llama-70B.}
\label{tab:appendix-deepseek-musique}
\resizebox{\linewidth}{!}{%
\begin{tabular}{lcccccc}
\toprule
System & 100 & 500 & 1k & 2k & 5k & 10k \\
\midrule
SAS & 0.294 & 0.383 & 0.407 & 0.418 & 0.419 & 0.417 \\
95\% CI & (0.282, 0.305) & (0.372, 0.397) & (0.400, 0.413) & (0.412, 0.429) & (0.413, 0.429) & (0.402, 0.429) \\
Avg. thinking tokens & 100 & 412 & 519 & 574 & 635 & 724 \\
Seq & 0.328 & 0.332 & 0.320 & 0.327 & 0.323 & 0.327 \\
95\% CI & (0.324, 0.347) & (0.317, 0.361) & (0.313, 0.336) & (0.312, 0.357) & (0.318, 0.344) & (0.320, 0.351) \\
Avg. thinking tokens & 99.6 & 479 & 788 & 1017 & 1105 & 1124 \\
SAS-L & 0.316 & 0.375 & 0.402 & 0.411 & 0.412 & 0.412 \\
95\% CI & (0.315, 0.335) & (0.362, 0.398) & (0.393, 0.428) & (0.400, 0.429) & (0.398, 0.434) & (0.398, 0.434) \\
Avg. thinking tokens & 100 & 459 & 623 & 704 & 755 & 827 \\
Debate & 0.308 & 0.320 & 0.315 & 0.352 & 0.357 & 0.360 \\
95\% CI & (0.304, 0.321) & (0.311, 0.342) & (0.299, 0.342) & (0.332, 0.364) & (0.354, 0.379) & (0.341, 0.368) \\
Avg. thinking tokens & 100 & 500 & 998 & 1805 & 2376 & 2506 \\
Ensemble & 0.316 & 0.319 & 0.323 & 0.346 & 0.334 & 0.330 \\
95\% CI & (0.311, 0.330) & (0.303, 0.324) & (0.317, 0.345) & (0.330, 0.371) & (0.325, 0.362) & (0.327, 0.337) \\
Avg. thinking tokens & 100 & 500 & 999 & 1868 & 2879 & 3267 \\
Subtask-parallel & 0.309 & 0.303 & 0.317 & 0.317 & 0.317 & 0.319 \\
95\% CI & (0.297, 0.327) & (0.297, 0.322) & (0.312, 0.337) & (0.315, 0.339) & (0.308, 0.338) & (0.309, 0.340) \\
Avg. thinking tokens & 100 & 492 & 884 & 1261 & 1484 & 1556 \\
Parallel-roles & 0.316 & 0.329 & 0.334 & 0.335 & 0.345 & 0.354 \\
95\% CI & (0.305, 0.326) & (0.323, 0.334) & (0.325, 0.349) & (0.317, 0.350) & (0.326, 0.362) & (0.341, 0.367) \\
Avg. thinking tokens & 100 & 500 & 996 & 1814 & 2647 & 2799 \\
\bottomrule
\end{tabular}}
\end{table*}

\subsection{Gemini 2.5 main results}
This subsection reports the Gemini results together with self-counted thought tokens, visible thought words, assumed visible thought tokens, and 95\% bootstrap confidence intervals (Table \ref{tab:appendix-gemflash-musique}, Table \ref{tab:appendix-gempro-musique}, Table \ref{tab:appendix-gemflash-frames}, and Table \ref{tab:appendix-gempro-frames}).

\begin{table*}[t]
\centering
\caption{MuSiQue 4-hop full results for Gemini-2.5-Flash.}
\label{tab:appendix-gemflash-musique}
\resizebox{\linewidth}{!}{%
\begin{tabular}{lcccccc}
\toprule
System & 100 & 500 & 1k & 2k & 5k & 10k \\
\midrule
SAS acc. & 0.263 & 0.340 & 0.331 & 0.334 & 0.344 & 0.338 \\
95\% CI & (0.249, 0.266) & (0.319, 0.358) & (0.324, 0.346) & (0.322, 0.353) & (0.323, 0.370) & (0.321, 0.361) \\
Self-count tok. & 90 & 399 & 637 & 884 & 1407 & 1687 \\
Visible words & 124 & 212 & 248 & 254 & 250 & 251 \\
Assumed visible tok. & 177 & 303 & 354 & 363 & 357 & 359 \\
\midrule
Sequential acc. & 0.272 & 0.287 & 0.287 & 0.283 & 0.289 & 0.285 \\
95\% CI & (0.258, 0.291) & (0.283, 0.309) & (0.270, 0.308) & (0.271, 0.296) & (0.273, 0.315) & (0.271, 0.291) \\
Self-count tok. & 84 & 366 & 599 & 885 & 1312 & 1681 \\
Visible words & 469 & 426 & 505 & 605 & 679 & 684 \\
Assumed visible tok. & 670 & 609 & 721 & 864 & 970 & 977 \\
\midrule
SAS-L acc. & 0.352 & 0.352 & 0.354 & 0.369 & 0.364 & 0.369 \\
95\% CI & (0.345, 0.355) & (0.334, 0.370) & (0.340, 0.358) & (0.356, 0.378) & (0.358, 0.375) & (0.358, 0.373) \\
Self-count tok. & 96 & 462 & 810 & 1443 & 2549 & 3000 \\
Visible words & 180 & 251 & 308 & 375 & 326 & 335 \\
Assumed visible tok. & 257 & 359 & 440 & 536 & 466 & 479 \\
\midrule
Subtask-parallel acc. & 0.264 & 0.291 & 0.313 & 0.323 & 0.332 & 0.335 \\
95\% CI & (0.257, 0.272) & (0.282, 0.310) & (0.308, 0.329) & (0.312, 0.345) & (0.317, 0.336) & (0.331, 0.357) \\
Self-count tok. & 74 & 373 & 723 & 1311 & 2387 & 3323 \\
Visible words & 437 & 440 & 582 & 751 & 929 & 963 \\
Assumed visible tok. & 624 & 629 & 831 & 1073 & 1327 & 1376 \\
\midrule
Parallel-roles acc. & 0.300 & 0.311 & 0.306 & 0.330 & 0.340 & 0.349 \\
95\% CI & (0.282, 0.322) & (0.297, 0.336) & (0.286, 0.326) & (0.300, 0.335) & (0.328, 0.352) & (0.331, 0.361) \\
Self-count tok. & 79 & 389 & 826 & 1689 & 3453 & 5147 \\
Visible words & 582 & 643 & 821 & 1114 & 1506 & 1603 \\
Assumed visible tok. & 831 & 919 & 1173 & 1591 & 2151 & 2290 \\
\midrule
Debate acc. & 0.318 & 0.329 & 0.328 & 0.349 & 0.355 & 0.354 \\
95\% CI & (0.297, 0.333) & (0.312, 0.346) & (0.312, 0.341) & (0.335, 0.361) & (0.346, 0.363) & (0.342, 0.364) \\
Self-count tok. & 73 & 392 & 833 & 1732 & 3294 & 4390 \\
Visible words & 512 & 491 & 654 & 912 & 1136 & 1180 \\
Assumed visible tok. & 731 & 701 & 934 & 1303 & 1623 & 1686 \\
\midrule
Ensemble acc. & 0.243 & 0.274 & 0.299 & 0.320 & 0.338 & 0.343 \\
95\% CI & (0.231, 0.251) & (0.268, 0.280) & (0.277, 0.315) & (0.310, 0.322) & (0.338, 0.344) & (0.340, 0.376) \\
Self-count tok. & 74 & 442 & 897 & 1742 & 3131 & 4363 \\
Visible words & 752 & 615 & 786 & 1014 & 1246 & 1295 \\
Assumed visible tok. & 1074 & 879 & 1123 & 1449 & 1780 & 1850 \\
\bottomrule
\end{tabular}}
\end{table*}

\begin{table*}[t]
\centering
\caption{MuSiQue 4-hop full results for Gemini-2.5-Pro.}
\label{tab:appendix-gempro-musique}
\resizebox{\linewidth}{!}{%
\begin{tabular}{lcccccc}
\toprule
System & 100 & 500 & 1k & 2k & 5k & 10k \\
\midrule
SAS acc. & 0.308 & 0.391 & 0.413 & 0.412 & 0.419 & 0.428 \\
95\% CI & (0.288, 0.320) & (0.369, 0.405) & (0.397, 0.420) & (0.408, 0.418) & (0.410, 0.425) & (0.413, 0.435) \\
Self-count tok. & 84 & 442 & 816 & 884 & 1526 & 1522 \\
Visible words & 124 & 226 & 273 & 254 & 276 & 275 \\
Assumed visible tok. & 177 & 323 & 390 & 363 & 394 & 393 \\
\midrule
Sequential acc. & 0.393 & 0.391 & 0.400 & 0.414 & 0.392 & 0.392 \\
95\% CI & (0.371, 0.404) & (0.379, 0.405) & (0.384, 0.414) & (0.396, 0.426) & (0.380, 0.410) & (0.380, 0.410) \\
Self-count tok. & 320 & 357 & 644 & 1388 & 2351 & 2351 \\
Visible words & 391 & 402 & 485 & 743 & 941 & 941 \\
Assumed visible tok. & 559 & 574 & 693 & 1061 & 1344 & 1344 \\
\midrule
SAS-L acc. & 0.373 & 0.423 & 0.414 & 0.449 & 0.455 & 0.436 \\
95\% CI & (0.345, 0.384) & (0.367, 0.438) & (0.383, 0.421) & (0.449, 0.461) & (0.438, 0.475) & (0.423, 0.449) \\
Self-count tok. & 77 & 403 & 907 & 1986 & 2260 & 2199 \\
Visible words & 146 & 269 & 378 & 355 & 348 & 351 \\
Assumed visible tok. & 209 & 384 & 540 & 507 & 497 & 501 \\
\midrule
Subtask-parallel acc. & 0.364 & 0.363 & 0.397 & 0.381 & 0.417 & 0.410 \\
95\% CI & (0.340, 0.391) & (0.355, 0.368) & (0.390, 0.417) & (0.368, 0.392) & (0.390, 0.428) & (0.405, 0.427) \\
Self-count tok. & 294 & 356 & 666 & 1573 & 3251 & 4003 \\
Visible words & 477 & 510 & 523 & 908 & 1004 & 1025 \\
Assumed visible tok. & 681 & 729 & 747 & 1297 & 1434 & 1464 \\
\midrule
Parallel-roles acc. & 0.400 & 0.397 & 0.430 & 0.409 & 0.447 & 0.447 \\
95\% CI & (0.380, 0.400) & (0.381, 0.400) & (0.416, 0.440) & (0.396, 0.412) & (0.441, 0.464) & (0.445, 0.461) \\
Self-count tok. & 349 & 349 & 627 & 1734 & 4301 & 6458 \\
Visible words & 626 & 653 & 658 & 1128 & 1326 & 1361 \\
Assumed visible tok. & 894 & 933 & 940 & 1611 & 1894 & 1944 \\
\midrule
Debate acc. & 0.412 & 0.435 & 0.444 & 0.448 & 0.470 & 0.458 \\
95\% CI & (0.392, 0.418) & (0.432, 0.462) & (0.430, 0.456) & (0.430, 0.476) & (0.453, 0.482) & (0.461, 0.477) \\
Self-count tok. & 310 & 310 & 661 & 1777 & 4644 & 6615 \\
Visible words & 464 & 384 & 532 & 888 & 1219 & 1223 \\
Assumed visible tok. & 663 & 549 & 760 & 1269 & 1741 & 1747 \\
\midrule
Ensemble acc. & 0.302 & 0.330 & 0.325 & 0.372 & 0.434 & 0.445 \\
95\% CI & (0.282, 0.309) & (0.319, 0.338) & (0.319, 0.331) & (0.357, 0.381) & (0.421, 0.435) & (0.436, 0.456) \\
Self-count tok. & 403 & 402 & 564 & 1362 & 3835 & 5764 \\
Visible words & 719 & 625 & 681 & 938 & 1361 & 1409 \\
Assumed visible tok. & 1027 & 893 & 973 & 1340 & 1944 & 2013 \\
\bottomrule
\end{tabular}}
\end{table*}

\begin{table*}[t]
\centering
\caption{FRAMES full results for Gemini-2.5-Flash.}
\label{tab:appendix-gemflash-frames}
\resizebox{\linewidth}{!}{%
\begin{tabular}{lcccccc}
\toprule
System & 100 & 500 & 1k & 2k & 5k & 10k \\
\midrule
SAS acc. & 0.333 & 0.487 & 0.551 & 0.532 & 0.545 & 0.547 \\
95\% CI & (0.315, 0.341) & (0.473, 0.496) & (0.537, 0.570) & (0.510, 0.537) & (0.541, 0.564) & (0.524, 0.551) \\
Self-count tok. & 89 & 374 & 535 & 714 & 1057 & 1267 \\
Visible words & 138 & 221 & 254 & 278 & 272 & 275 \\
Assumed visible tok. & 197 & 316 & 363 & 397 & 389 & 393 \\
\midrule
Sequential acc. & 0.462 & 0.494 & 0.507 & 0.526 & 0.524 & 0.516 \\
95\% CI & (0.446, 0.477) & (0.471, 0.515) & (0.482, 0.530) & (0.489, 0.544) & (0.503, 0.550) & (0.509, 0.550) \\
Self-count tok. & 73 & 334 & 517 & 715 & 984 & 1223 \\
Visible words & 527 & 395 & 458 & 521 & 567 & 569 \\
Assumed visible tok. & 753 & 564 & 654 & 744 & 810 & 813 \\
\midrule
SAS-L acc. & 0.427 & 0.459 & 0.484 & 0.517 & 0.542 & 0.546 \\
95\% CI & (0.421, 0.446) & (0.438, 0.472) & (0.466, 0.497) & (0.500, 0.542) & (0.516, 0.548) & (0.520, 0.560) \\
Self-count tok. & 94 & 452 & 810 & 1299 & 1970 & 2158 \\
Visible words & 237 & 301 & 383 & 411 & 400 & 389 \\
Assumed visible tok. & 339 & 430 & 547 & 587 & 571 & 556 \\
\midrule
Debate acc. & 0.445 & 0.470 & 0.527 & 0.527 & 0.552 & 0.564 \\
95\% CI & (0.428, 0.489) & (0.458, 0.471) & (0.503, 0.549) & (0.505, 0.544) & (0.542, 0.557) & (0.545, 0.574) \\
Self-count tok. & 79 & 404 & 847 & 1580 & 2679 & 3193 \\
Visible words & 598 & 570 & 716 & 941 & 1137 & 1135 \\
Assumed visible tok. & 854 & 814 & 1023 & 1344 & 1624 & 1621 \\
\midrule
Ensemble acc. & 0.284 & 0.384 & 0.458 & 0.498 & 0.539 & 0.559 \\
95\% CI & (0.276, 0.301) & (0.383, 0.400) & (0.444, 0.470) & (0.492, 0.513) & (0.527, 0.551) & (0.538, 0.560) \\
Self-count tok. & 76 & 450 & 889 & 1556 & 2465 & 3191 \\
Visible words & 874 & 668 & 770 & 954 & 1116 & 1157 \\
Assumed visible tok. & 1249 & 954 & 1100 & 1363 & 1594 & 1653 \\
\midrule
Subtask-parallel acc. & 0.365 & 0.442 & 0.482 & 0.509 & 0.532 & 0.541 \\
95\% CI & (0.333, 0.382) & (0.429, 0.466) & (0.472, 0.491) & (0.491, 0.527) & (0.518, 0.542) & (0.530, 0.553) \\
Self-count tok. & 68 & 354 & 644 & 1048 & 1775 & 2320 \\
Visible words & 420 & 405 & 511 & 622 & 732 & 746 \\
Assumed visible tok. & 600 & 579 & 730 & 889 & 1046 & 1066 \\
\midrule
Parallel-roles acc. & 0.441 & 0.451 & 0.495 & 0.536 & 0.545 & 0.556 \\
95\% CI & (0.405, 0.448) & (0.424, 0.453) & (0.481, 0.509) & (0.514, 0.545) & (0.534, 0.555) & (0.537, 0.559) \\
Self-count tok. & 81 & 395 & 816 & 1505 & 2803 & 3755 \\
Visible words & 673 & 718 & 819 & 1021 & 1282 & 1347 \\
Assumed visible tok. & 961 & 1026 & 1170 & 1459 & 1831 & 1924 \\
\bottomrule
\end{tabular}}
\end{table*}

\begin{table*}[t]
\centering
\caption{FRAMES full results for Gemini-2.5-Pro.}
\label{tab:appendix-gempro-frames}
\resizebox{\linewidth}{!}{%
\begin{tabular}{lcccccc}
\toprule
System & 100 & 500 & 1k & 2k & 5k & 10k \\
\midrule
SAS acc. & 0.368 & 0.600 & 0.680 & 0.700 & 0.700 & 0.692 \\
95\% CI & (0.358, 0.376) & (0.578, 0.608) & (0.640, 0.704) & (0.673, 0.705) & (0.691, 0.705) & (0.661, 0.718) \\
Self-count tok. & 84 & 494 & 807 & 1119 & 1158 & 1210 \\
Visible words & 138 & 261 & 322 & 337 & 337 & 330 \\
Assumed visible tok. & 197 & 373 & 460 & 481 & 481 & 471 \\
\midrule
Sequential acc. & 0.654 & 0.660 & 0.670 & 0.690 & 0.680 & 0.691 \\
95\% CI & (0.639, 0.687) & (0.640, 0.706) & (0.652, 0.705) & (0.651, 0.730) & (0.671, 0.693) & (0.660, 0.708) \\
Self-count tok. & 281 & 354 & 683 & 1312 & 2014 & 2152 \\
Visible words & 419 & 434 & 533 & 766 & 919 & 923 \\
Assumed visible tok. & 599 & 620 & 761 & 1094 & 1313 & 1319 \\
\midrule
SAS-L acc. & 0.451 & 0.450 & 0.610 & 0.680 & 0.690 & 0.692 \\
95\% CI & (0.435, 0.473) & (0.444, 0.463) & (0.589, 0.619) & (0.665, 0.697) & (0.670, 0.700) & (0.664, 0.704) \\
Self-count tok. & 80 & 390 & 902 & 1541 & 1626 & 1591 \\
Visible words & 148 & 315 & 428 & 418 & 348 & 414 \\
Assumed visible tok. & 211 & 450 & 611 & 597 & 497 & 591 \\
\midrule
Debate acc. & 0.649 & 0.660 & 0.638 & 0.660 & 0.700 & 0.697 \\
95\% CI & (0.632, 0.658) & (0.632, 0.676) & (0.615, 0.663) & (0.624, 0.669) & (0.687, 0.720) & (0.670, 0.712) \\
Self-count tok. & 317 & 315 & 673 & 1854 & 4371 & 5431 \\
Visible words & 482 & 481 & 625 & 1032 & 1392 & 1430 \\
Assumed visible tok. & 689 & 687 & 893 & 1474 & 1989 & 2043 \\
\midrule
Ensemble acc. & 0.407 & 0.400 & 0.430 & 0.558 & 0.710 & 0.719 \\
95\% CI & (0.400, 0.420) & (0.395, 0.436) & (0.411, 0.447) & (0.546, 0.564) & (0.694, 0.712) & (0.704, 0.750) \\
Self-count tok. & 406 & 408 & 587 & 1426 & 3552 & 4543 \\
Visible words & 670 & 673 & 737 & 1011 & 1411 & 1452 \\
Assumed visible tok. & 957 & 961 & 1053 & 1444 & 2016 & 2074 \\
\midrule
Subtask-parallel acc. & 0.530 & 0.562 & 0.596 & 0.637 & 0.652 & 0.655 \\
95\% CI & (0.499, 0.555) & (0.547, 0.591) & (0.583, 0.638) & (0.625, 0.660) & (0.634, 0.676) & (0.637, 0.677) \\
Self-count tok. & 219 & 350 & 772 & 1481 & 2414 & 2667 \\
Visible words & 355 & 409 & 558 & 772 & 926 & 915 \\
Assumed visible tok. & 507 & 584 & 797 & 1103 & 1323 & 1307 \\
\midrule
Parallel-roles acc. & 0.609 & 0.598 & 0.600 & 0.658 & 0.700 & 0.718 \\
95\% CI & (0.597, 0.633) & (0.583, 0.633) & (0.593, 0.625) & (0.648, 0.676) & (0.696, 0.723) & (0.711, 0.736) \\
Self-count tok. & 343 & 344 & 640 & 1789 & 3871 & 5085 \\
Visible words & 619 & 630 & 739 & 1099 & 1514 & 1558 \\
Assumed visible tok. & 884 & 900 & 1056 & 1570 & 2163 & 2226 \\
\bottomrule
\end{tabular}}
\end{table*}

\subsection{Paraphrase results}
This subsection reports the paraphrase results with confidence intervals for the original, light paraphrased, and deep paraphrased settings (Table \ref{tab:appendix-paraphrase}).

\begin{table*}[t]
\centering
\caption{Paraphrase results with 95\% bootstrap confidence intervals.}
\label{tab:appendix-paraphrase}
\resizebox{\linewidth}{!}{%
\begin{tabular}{ll|cc|cc|cc}
\toprule
& & \multicolumn{2}{c|}{Original} & \multicolumn{2}{c|}{Light paraphrase} & \multicolumn{2}{c}{Deep paraphrase} \\
\cmidrule(lr){3-4}\cmidrule(lr){5-6}\cmidrule(lr){7-8}
Model & Budget & SAS & Seq & SAS & Seq & SAS & Seq \\
\midrule
Qwen3-30B-A3B & 1k & 0.260 (0.260, 0.271) & 0.229 (0.223, 0.248) & 0.249 (0.236, 0.260) & 0.224 (0.175, 0.234) & 0.268 (0.264, 0.286) & 0.225 (0.214, 0.234) \\
Qwen3-30B-A3B & 2k & 0.271 (0.269, 0.276) & 0.229 (0.223, 0.240) & 0.264 (0.254, 0.274) & 0.204 (0.200, 0.220) & 0.272 (0.263, 0.286) & 0.236 (0.231, 0.248) \\
Gemini-2.5-Flash & 1k & 0.331 (0.324, 0.346) & 0.287 (0.270, 0.308) & 0.326 (0.320, 0.331) & 0.278 (0.266, 0.298) & 0.358 (0.346, 0.373) & 0.308 (0.283, 0.325) \\
Gemini-2.5-Flash & 2k & 0.334 (0.322, 0.353) & 0.283 (0.271, 0.296) & 0.329 (0.323, 0.340) & 0.289 (0.271, 0.304) & 0.348 (0.332, 0.356) & 0.304 (0.277, 0.309) \\
\bottomrule
\end{tabular}}
\end{table*}

\subsection{Multiple Gemini model version results}
This subsection reports the unlimited thinking results for multiple different Gemini model versions with confidence intervals (Table \ref{tab:appendix-gemini-versions}).

\begin{table}[t]
\centering
\caption{MuSiQue 4-hop accuracy across Gemini model versions with unlimited thinking, with 95\% bootstrap confidence intervals.}
\label{tab:appendix-gemini-versions}
\begin{tabular}{lcc}
\toprule
Model & SAS & Sequential \\
\midrule
Gemini-2-Flash-Lite & 0.258 (0.245, 0.282) & 0.255 (0.226, 0.266) \\
Gemini-2-Flash & 0.347 (0.333, 0.366) & 0.308 (0.296, 0.321) \\
Gemini-2.5-Flash & 0.296 (0.276, 0.319) & 0.294 (0.279, 0.316) \\
Gemini-2.5-Pro & 0.399 (0.381, 0.400) & 0.396 (0.372, 0.403) \\
Gemini-3-Pro-Preview & 0.491 (0.472, 0.500) & 0.489 (0.461, 0.513) \\
\bottomrule
\end{tabular}
\end{table}

\subsection{Context degradation results}
This subsection reports the full context degradation results for all four perturbation families together with their 95\% bootstrap confidence intervals (Table \ref{tab:appendix-context-degradation}).

\begin{table*}[t]
\centering
\caption{Context-degradation results on MuSiQue 4-hop with Qwen3-30B-A3B and a fixed 1000-token thinking budget.}
\label{tab:appendix-context-degradation}
%\resizebox{\linewidth}{!}{%
\begin{tabular}{llcc}
\toprule
Method & Level & SAS & Sequential \\
\midrule
Deletion & $\alpha = 0.3$ & 0.245 (0.234, 0.248) & 0.221 (0.218, 0.229) \\
Deletion & $\alpha = 0.5$ & 0.216 (0.210, 0.226) & 0.219 (0.217, 0.235) \\
Deletion & $\alpha = 0.7$ & 0.229 (0.221, 0.242) & 0.215 (0.213, 0.225) \\
\midrule
Masking & $\alpha = 0.3$ & 0.234 (0.223, 0.255) & 0.221 (0.211, 0.229) \\
Masking & $\alpha = 0.5$ & 0.223 (0.205, 0.238) & 0.223 (0.206, 0.231) \\
Masking & $\alpha = 0.7$ & 0.220 (0.215, 0.224) & 0.231 (0.224, 0.248) \\
\midrule
Substitution & $\alpha = 0.3$ & 0.243 (0.232, 0.253) & 0.217 (0.206, 0.226) \\
Substitution & $\alpha = 0.5$ & 0.231 (0.215, 0.241) & 0.223 (0.216, 0.235) \\
Substitution & $\alpha = 0.7$ & 0.200 (0.185, 0.211) & 0.225 (0.220, 0.240) \\
\midrule
Distractors & $k = 10$ & 0.257 (0.256, 0.269) & 0.228 (0.219, 0.248) \\
Distractors & $k = 20$ & 0.254 (0.250, 0.262) & 0.232 (0.230, 0.239) \\
Distractors & $k = 30$ & 0.256 (0.243, 0.257) & 0.233 (0.219, 0.241) \\
\bottomrule
\end{tabular}
%}
\end{table*}

\section{Diagnostic: On Gemini Thought Token Accounting}
\label{appendix:gemini_accounting}

As introduced in the paper, we use the official \texttt{thinkingBudget} parameter for Gemini models. However, official documentation \footnote{https://ai.google.dev/gemini-api/docs/thinking} defines this as a \emph{guide} rather than a hard cap and discloses model-specific min/max ranges and possible over/underflow. This aligns with community reports of mismatches between requested budget and observed output \footnote{This is mentioned in https://github.com/googleapis/python-genai/issues/782}.

Our own experiments confirm a significant and complex discrepancy between the API-reported \texttt{thoughts\_token\_count} and the actual token count of the visible thought content. We tracked three values for all Gemini experiments: (i) the API-reported token count, (ii) a content-based word count, and (iii) a proxy token count derived by multiplying the word count by $10/7$ \footnote{This number is obtained from averaging the word-to-token ratio mentioned in https://ai.google.dev/gemini-api/docs/tokens}.

Our analysis reveals three key findings:

\begin{itemize}
    \item \textbf{API-Reported counts are highly inflated at higher thinking budget.} The \texttt{thoughts\_token\_count} returned by the API is often substantially larger than the token count of the visible text at higher thinking budget. For example, using Gemini-2.5-Flash with a \emph{requested} 10k budget, the SAS pipeline reported an average of 1,687 thinking tokens. However, the visible thought content only contained an average of 251 words, a proxy for 359 tokens (see Table \ref{tab:appendix-gemflash-musique}). This represents a 4.7x inflation factor, suggesting the API may be counting internal reasoning steps that are never externalized in the text.

    \item \textbf{Visible thought content plateaus.} We observe that for the standard SAS prompt and Sequential MAS, the \emph{actual length} of the visible thought text (our proxy) hits a ceiling and stops growing, even as the requested budget increases. For Gemini-2.5-Flash SAS, the visible thought proxy plateaus at $\approx$350 tokens. It is 354 tokens at a 1k budget and 359 tokens at a 10k budget (see Table \ref{tab:appendix-gemflash-musique}). This suggests that simply increasing the \texttt{thinkingBudget} parameter does not guarantee more extensive reasoning, even when the API reports so. Our SAS with longer thinking variant successfully produced more visible text (e.g., 479 tokens at 10k), confirming that prompt-level incentives are also critical.
    
    \item \textbf{Sequential MAS produces more visible thought text than SAS.} At a matched \emph{requested} budget, the Sequential MAS pipeline consistently produced more total visible thought text than the SAS pipeline. At a 1k budget on Pro, MAS produced 693 proxy tokens vs. SAS's 390 (see Table \ref{tab:appendix-gempro-musique}). This is likely because the Sequential MAS pipeline executes $k$ separate agent calls, and the final thought text is a concatenation of $k$ distinct (and potentially truncated) thought blocks, leading to more total text output.
\end{itemize}

These accounting artifacts make a direct ``apples-to-apples'' compute comparison based on \emph{observed} tokens (either API-reported or content-based) intractable. The API-reported count is unreliable, and the proxy visible token is non-linear with the requested budget.

Therefore, our primary analysis \textbf{intentionally and necessarily compares configurations based on their matched \emph{requested} budget $B$}. This is the only variable that can be directly controlled by the researcher. We argue this remains the most fair and reproducible method for evaluation, even if the underlying models utilize that budget in different and opaque ways. We strongly recommend researchers to acknowledge this discrepancy and call on API providers to clarify how \texttt{thoughts\_token\_count} is calculated.

\section{Gemini Results across different version:}
Table \ref{tab:appendix-gemini-versions} shows details results of Gemini model across different version

\end{document}